\documentclass[12pt]{article}
\usepackage{amsfonts}
\usepackage{amsmath}
\usepackage{amssymb}
\usepackage{color}
\usepackage{graphicx}
\usepackage{mathrsfs} % for math script
\usepackage{natbib}
\usepackage{pstricks, pst-node}

\newtheorem{lemma}{Lemma}
\newtheorem{theorem}{Theorem}

\def\sign{\hbox{\rm sgn}}
\def\polylog{{\rm polylog}}
\def\rank{{\rm rank}}

\def\lam{\lambda}
\def\real{{\mathbb{R}}}
\def\R{{\real}}

\newcommand{\bel}{\begin{eqnarray}\label}
\newcommand{\eel}{\end{eqnarray}}
\newcommand{\bes}{\begin{eqnarray*}}
\newcommand{\ees}{\end{eqnarray*}}

\def\benu{\begin{enumerate}}
\def\eenu{\end{enumerate}}

\def\argmax{\mathop{\rm arg\, max}}

\def\real{{\mathbb{R}}}
\def\R{{\real}}
\def\E{{\mathbb{E}}}
\def\P{{\mathbb{P}}}

\def\complex{\mathop{{\rm I}\kern-.58em\hbox{\rm C}}\nolimits}

\def\rank{\hbox{\rm rank}}

\def\sgn{\hbox{\rm sgn}}
\def\trace{\hbox{\rm trace}}

\def\supp{\hbox{\rm supp}}

\def\mathbold{\boldsymbol} %\def\mathbold{\mathbf}
\def\ba{\mathbold{a}}

\def\Atil{{\widetilde A}}

\def\bb{\mathbold{b}}

\def\scrB{{\mathscr B}}\def\Btil{{\widetilde B}}

\def\Ctil{{\widetilde C}}

\def\bc{\mathbold{c}}

\def\Ctil{{\widetilde C}}

\def\calD{{\cal D}}\def\scrD{{\mathscr D}}\def\Dtil{{\widetilde D}}

\def\bfe{\mathbold{e}}

\def\bG{\mathbold{G}}

\def\bI{\mathbold{I}}
\def\calI{{\cal I}}

\def\calL{{\cal L}}

\def\bP{\mathbold{P}}
\def\calP{{\cal P}}

\def\calQ{{\cal Q}}

\def\rbar{{\overline r}}

\def\calR{{\cal R}}

\def\bT{\mathbold{T}}\def\hbT{{\widehat{\bT}}}

\def\bu{\mathbold{u}}

\def\bU{\mathbold{U}}
\def\scrU{{\mathscr U}}

\def\bv{\mathbold{v}}

\def\bw{\mathbold{w}}

\def\bW{\mathbold{W}}

\def\bX{\mathbold{X}}

\def\bY{\mathbold{Y}}

\def\bZ{\mathbold{Z}}

\def\talpha{\widetilde{\alpha}}

\def\bDelta{\mathbold{\Delta}}

\def\eps{\epsilon}

\def\lam{\lambda}

\begin{document}

\title{On Tensor Completion via Nuclear Norm Minimization}
\author{Ming Yuan$^\ast$ ~and~ Cun-Hui Zhang$^\dag$\\
University of Wisconsin-Madison and Rutgers University}

\footnotetext[1]{
Department of Statistics, University of Wisconsin-Madison, Madison, WI 53706. The research of Ming Yuan was supported in part by NSF Career Award DMS-1321692 and FRG Grant DMS-1265202.}
\footnotetext[2]{
Department of Statistics and Biostatistics, Rutgers University, Piscataway, New Jersey 08854. The research of Cun-Hui Zhang was supported in part by NSF Grants DMS-1129626 and DMS-1209014}

\date{(\today)}

\maketitle

\begin{abstract}
Many problems can be formulated as recovering a low-rank tensor. Although an increasingly common task, tensor recovery remains a challenging problem because of the delicacy associated with the decomposition of higher order tensors. To overcome these difficulties, existing approaches often proceed by unfolding tensors into matrices and then apply techniques for matrix completion. We show here that such matricization fails to exploit the tensor structure and may lead to suboptimal procedure. More specifically, we investigate a convex optimization approach to tensor completion by directly minimizing a tensor nuclear norm and prove that this leads to an improved sample size requirement. To establish our results, we develop a series of algebraic and probabilistic techniques such as characterization of subdifferetial for tensor nuclear norm and concentration inequalities for tensor martingales, which may be of independent interests and could be useful in other tensor related problems.
\end{abstract}
\newpage

\section{Introduction}
Let $\bT\in \R^{d_1\times d_2\times\cdots\times d_N}$ be an $N$th order tensor, and $\Omega$ be a randomly sampled subset of $[d_1]\times\cdots\times [d_N]$ where $[d]=\{1,2,\ldots, d\}$. The goal of tensor completion is to recover $\bT$ when observing only entries $\bT(\omega)$ for $\omega\in \Omega$. In particular, we are interested in the case when the dimensions $d_1,\ldots, d_N$ are large. Such a problem arises naturally in many applications. Examples include hyper-spectral image analysis (Li and Li, 2010), multi-energy computed tomography (Semerci et al., 2013), radar signal processing (Sidiropoulos and Nion, 2010), audio classification (Mesgarani, Slaney and Shamma, 2006) and text mining (Cohen and Collins, 2012) among numerous others. Common to these and many other problems, the tensor $\bT$ can oftentimes be identified with a certain low-rank structure. The low-rankness entails reduction in degrees of freedom, and as a result, it is possible to recover $\bT$ exactly even when the sample size $|\Omega|$ is much smaller than the total number, $d_1d_2\cdots d_N$, of entries in $\bT$.

In particular, when $N=2$, this becomes the so-called matrix completion problem which has received considerable amount of attention in recent years. See, e.g., Cand\`es and Recht (2008), Cand\`es and Tao (2009), Recht (2010), and Gross (2011) among many others. An especially attractive approach is through nuclear norm minimization:
$$
\min_{\bX\in \R^{d_1\times d_2}} \|\bX\|_\ast\qquad {\rm subject\ to\ }\bX(\omega)=\bT(\omega)\quad \forall \omega\in \Omega,
$$
where the nuclear norm $\|\cdot\|_\ast$ of a matrix is given by
$$
\|\bX\|_\ast=\sum_{k=1}^{\min\{d_1,d_2\}}\sigma_k(\bX),
$$
and $\sigma_k(\cdot)$ stands for the $k$th largest singular value of a matrix. Denote by $\widehat{\bT}$ the solution to the aforementioned nuclear norm minimization problem. As shown, for example, by Gross (2011), if an unknown $d_1\times d_2$ matrix $\bT$ of rank $r$ is of low coherence with respect to the canonical basis, then it can be perfectly reconstructed by $\widehat{\bT}$ with high probability whenever $|\Omega|\ge C (d_1+d_2)r\log^2(d_1+d_2)$, where $C$ is a numerical constant. In other words, perfect recovery of a matrix is possible with observations from a very small fraction of entries in $\bT$.

In many practical situations, we need to consider higher order tensors. The seemingly innocent task of generalizing these ideas from matrices to higher order tensor completion problems, however, turns out to be rather subtle, as basic notion such as rank, or singular value decomposition, becomes ambiguous for higher order tensors (e.g., Kolda and Bader, 2009; Hillar and Lim, 2013). A common strategy to overcome the challenges in dealing with high order tensors is to unfold them to matrices, and then resort to the usual nuclear norm minimization heuristics for matrices. To fix ideas, we shall focus on third order tensors ($N=3$) in the rest of the paper although our techniques can be readily used to treat higher order tensor. %with the exception of Section \ref{sec:disc} where we briefly discuss the implications of our results on higher order tensors. 
%%%
Following the matricization approach, 
$\bT$ can be reconstructed by the solution of the following convex program:
$$
\min_{\bX\in \R^{d_1\times d_2\times d_3}} \{\|X^{(1)}\|_\ast+\|X^{(2)}\|_\ast+\|X^{(3)}\|_\ast\}\qquad {\rm subject\ to\ }\bX(\omega)=\bT(\omega)\quad \forall \omega\in \Omega,
$$
where $X^{(j)}$ is a $d_j\times (\prod_{k\neq j}d_k)$ matrix whose columns are the mode-$j$ fibers of $\bX$. See, e.g., Liu et al. (2009), Signoretto, Lathauwer and Suykens (2010), Gandy et al. (2011), Tomioka, Hayashi and Kashima (2010), and Tomioka et al. (2011). In the light of existing results on matrix completion, with this approach, $\bT$ can be reconstructed perfectly with high probability provided that
$$|\Omega|\ge C (d_1d_2r_3+d_1r_2d_3+r_1d_2d_3)\log^2(d_1+d_2+d_3)$$ 
uniformly sampled entries are observed, where $r_j$ is the rank of $X^{(j)}$ and $C$ is a numerical constant. See, e.g., Mu et al. (2013). It is of great interests to investigate if this sample size requirement can be improved by avoiding matricization of tensors. We show here that the answer indeed is affirmative and a more direct nuclear norm minimization formulation requires a smaller sample size to recover $\bT$.

More specifically, write, for two tensors $\bX, \bY\in \R^{d_1\times d_2\times d_3}$,
$$
\langle \bX, \bY\rangle =\sum_{\omega\in [d_1]\times[d_2]\times [d_3]} \bX(\omega)\bY(\omega)
$$
as their inner product. Define
\bes
\|\bX\|=\max_{\substack{\bu_j\in \R^{d_j}: \|\bu_1\|=\|\bu_2\|=\|\bu_3\|=1}}\langle\bX, \bu_1\otimes\bu_2\otimes \bu_3\rangle,
\ees
where, with slight abuse of notation, $\|\cdot\|$ also stands for the usual Euclidean norm for a vector, and for vectors $\bu_j=(u_1^j,\ldots,u_{d_j}^j)^\top$,
$$
\bu_1\otimes\bu_2\otimes \bu_3=(u_a^1u_b^2u_c^3)_{1\le a\le d_1,1\le b\le d_2,1\le c\le d_3}.
$$
It is clear that the $\|\cdot\|$ defined above for tensors is a norm and can be viewed as an extension of the usual matrix spectral norm. Appealing to the duality between the spectral norm and nuclear norm in the matrix case, we now consider the following nuclear norm for tensors:
\bes
\|\bX\|_\ast=\max_{\bY\in \R^{d_1\times d_2\times d_3}: \|\bY\|\le 1}\langle \bY, \bX\rangle.
\ees
It is clear that $\|\cdot\|_\ast$ is also a norm. We then consider reconstructing $\bT$ via the solution to the following convex program:
$$
\min_{\bX\in\R^{d_1\times d_2\times d_3}} \|\bX\|_\ast\qquad {\rm subject\ to\ } \bX(\omega)=\bT(\omega)\quad \forall \omega\in \Omega.
$$
We show that the sample size requirement for perfect recovery of a tensor with low coherence using this approach is
$$|\Omega|\ge C\left(r^2(d_1+d_2+d_3)+\sqrt{rd_1d_2d_3}\right)\polylog\left(d_1+d_2+d_3\right),$$
where 
$$
r=\sqrt{(r_1r_2d_3+r_1r_3d_2+r_2r_3d_1)/(d_1+d_2+d_3)},
$$
$\polylog(x)$ is a certain polynomial function of $\log(x)$, and $C$ is a numerical constant. In particular, when considering (nearly) cubic tensors with $d_1,d_2$ and $d_3$ approximately equal to a common $d$, then this sample size requirement is essentially of the order $r^{1/2}(d\log d)^{3/2}$. In the case when the tensor dimension $d$ is large while the rank $r$ is relatively small, this can be a drastic improvement over the existing results based on matricizing tensors where the sample size requirement is $r(d\log d)^2$.

The high-level strategy to the investigation of the proposed nuclear norm minimization approach for tensors is similar, in a sense, to the treatment of matrix completion. Yet the analysis for tensors is much more delicate and poses significant new challenges because many of the well-established tools for matrices, either algebraic such as characterization of 
%%%
the subdifferential of the nuclear norm, 
or probabilistic such as concentration inequalities for martingales, do not exist for tensors. Some of these disparities can be bridged and we develop various tools to do so. Others are due to fundamental differences between matrices and higher order tensors, and we devise new strategies to overcome them. The tools and techniques we developed may be of independent interests and can be useful in dealing with other problems for tensors.

The rest of the paper is organized as follows. We first describe some basic properties of tensors 
%%%
and their nuclear norm necessary for our analysis in Section \ref{sec:notation}. 
Section \ref{sec:main} discusses the main architect of our analysis. The main probabilistic tools we use are concentration bounds for the sum of random tensors. Because the tensor spectral norm does not have the interpretation as an operator norm of a linear mapping between Hilbert spaces, the usual matrix Bernstein inequality cannot be directly applied. It turns out that different strategies are required for tensors of low rank and tensors with sparse support, and these results are presented in Sections \ref{sec:low} and \ref{sec:sparse} respectively. We conclude the paper with a few remarks in Section \ref{sec:disc}.

\section{Tensor}
\label{sec:notation}

We first collect some useful algebraic facts for tensors essential to our later analysis.
%
%\subsection{Norms}
Recall that the inner product between two third order tensors $\bX, \bY\in \R^{d_1\times d_2\times d_3}$ is given by
$$
\langle \bX, \bY\rangle =\sum_{a=1}^{d_1}\sum_{b=1}^{d_2}\sum_{c=1}^{d_3} \bX(a,b,c)\bY(a,b,c),
$$
and $\|\bX\|_{\rm HS}=\langle \bX, \bX\rangle^{1/2}$ is the usual Hilbert-Schmidt norm of $\bX$. Another tensor norm of interest is the entrywise $\ell_\infty$ norm, or tensor max norm:
$$\|\bX\|_{\max}=\max_{\omega\in [d_1]\times[d_2]\times[d_3]}|\bX(\omega)|.$$
It is clear that
%\begin{lemma}\label{prop-1}
for any the third order tensor $\bX\in \R^{d_1\times d_2\times d_3}$, 
$$
\|\bX\|_{\max}\le \|\bX\|\le\|\bX\|_{\rm HS}\le \|\bX\|_*,\quad {\rm and}\quad
\|\bX\|_{\rm HS}^2\le \|\bX\|_*\|\bX\|. 
$$
%\end{lemma}

We shall also encounter linear maps defined on tensors. Let $\calR: \R^{d_1\times d_2\times d_3}\to \R^{d_1\times d_2\times d_3}$ be a linear map. We define the induced operator norm of $\calR$ under tensor Hilbert-Schmidt norm as
$$
\|\calR\|=\max\Big\{\|\calR\bX\|_{\rm HS}: \bX\in\R^{d_1\times d_2\times d_3}, \|\bX\|_{\rm HS}\le 1\Big\}.
$$
%and
%$$
%\|\calR\|_{{\max}\to {\max}}=\max\Big\{\|\calR\bX\|_{\max}: \bX\in\R^{d_1\times d_2\times d_3}, \|\bX\|_{\max}\le 1\Big\}.
%$$
%For brevity, we shall omit the subscript ${\rm HS}\to {\rm HS}$ when referring to the operator norm under tensor Hilbert-Schmidt norms.

\subsection{Decomposition and Projection}
Consider the following tensor decomposition of $\bX$ into rank-one tensors:
\bel{eq:tensor}
\bX=[A, B, C]:=\sum_{k=1}^r\ba_k\otimes \bb_k\otimes\bc_k,
\eel
where $\ba_k$s, $\bb_k$s and $\bc_k$s are the column vectors of matrices $A$, $B$ and $C$ respectively. Such a decomposition in general is not unique (see, e.g., Kruskal, 1989). However, the linear spaces spanned by columns of $A$, $B$ and $C$ respectively are uniquely defined.

%%%
More specifically, write $\bX(\cdot,b,c)=(\bX(1,b,c),\ldots,\bX(d_1,b,c))^{\top}$, that is the mode-1 fiber of $\bX$. Define $\bX(a,\cdot,c)$ and $\bX(a,b,\cdot)$ in a similar fashion. Let
\bes
\calL_1(\bX)&=&{\rm l.s.}\{\bX(\cdot,b,c): 1\le b\le d_2, 1\le c\le d_3\};\\
\calL_2(\bX)&=&{\rm l.s.}\{\bX(a,\cdot,c): 1\le a\le d_1, 1\le c\le d_3\};\\
\calL_3(\bX)&=&{\rm l.s.}\{\bX(a,b,\cdot): 1\le a\le d_1, 1\le b\le d_2\},
\ees
where ${\rm l.s.}$ represents the linear space spanned by a collection of vectors of conformable dimension. Then it is clear that the linear space spanned by the column vectors of $A$ is $\calL_1(\bX)$, and similar statements hold true for the column vectors of $B$ and $C$. In the case of matrices, both marginal linear spaces, $\calL_1$ and $\calL_2$ are necessarily of the same dimension as they are spanned by the respective singular vectors. For higher order tensors, however, this is typically not true. We shall denote by $r_j(\bX)$ the dimension of $\calL_j(\bX)$ for $j=1,2$ and $3$, which are often referred to the Tucker ranks of $\bX$. Another useful notion of ``tensor rank'' for our purposes is
$$
\rbar(\bX)=\sqrt{\left(r_1(\bX)r_2(\bX)d_3+r_1(\bX)r_3(\bX)d_2+r_2(\bX)r_3(\bX)d_1\right)/d}.
$$
where $d=d_1+d_2+d_3$, which can also be viewed as a generalization of the matrix rank to tensors. It is well known that the smallest value for $r$ in the rank-one decomposition (\ref{eq:tensor}) is in $[\rbar(\bX), \rbar^2(\bX)]$.

Let $M$ be a matrix of size $d_0\times d_1$. Marginal multiplication of $M$ and a tensor $\bX$ in the first coordinate yields a tensor of size $d_0\times d_2\times d_3$:
$$
(M\times_1 \bX)(a,b,c)=\sum_{a'=1}^{d_1} M_{aa'}X(a',b,c).
$$
It is easy to see that if $\bX=[A, B, C]$, then $M\times_1 \bX=[MA, B, C]$. Marginal multiplications $\times_2$ and $\times_3$ between a matrix of conformable size and $\bX$ can be similarly defined.

Let $\bP$ be arbitrary projection from $\R^{d_1}$ to a linear subspace of $\R^{d_1}$. It is clear from the definition of marginal multiplications, $[\bP A, B, C]$ is also uniquely defined for tensor $\bX=[A, B, C]$, that is, $[\bP A, B, C]$ does not depend on the particular decomposition of $A$, $B$, $C$. Now let $\bP_j$ be arbitrary projection from $\R^{d_j}$ to a linear subspace of $\R^{d_j}$. Define a tensor projection $\bP_1\otimes\bP_2\otimes \bP_3$ on $\bX=[A,B,C]$ as
$$(\bP_1\otimes\bP_2\otimes \bP_3)\bX=[\bP_1A, \bP_2B, \bP_3C].$$
We note that there is no ambiguity in defining $(\bP_1\otimes\bP_2\otimes \bP_3)\bX$ because of the uniqueness of marginal projections.

Recall that $\calL_j(\bX)$ is the linear space spanned by the mode-$j$ fibers of $\bX$. Let $\bP_{\bX}^j$ be the projection from $\R^{d_j}$ to $\calL_j(\bX)$, and $\bP_{\bX^\perp}^j$ be the projection to its orthogonal complement in $\R^{d_j}$. The following tensor projections will be used extensively in our analysis:
\bes
\calQ_{\bX}^0 =  \bP_{\bX}^1\otimes \bP_{\bX}^2 \otimes \bP_{\bX}^3, && \calQ_{\bX^\perp}^0 =  \bP_{\bX^\perp}^1\otimes \bP_{\bX^\perp}^2 \otimes \bP_{\bX^\perp}^3, \cr
\calQ_{\bX}^1 =  \bP_{\bX^\perp}^1\otimes \bP_{\bX}^2\otimes \bP_{\bX}^3, && \calQ_{\bX^\perp}^1 =  \bP_{\bX}^1\otimes \bP_{\bX^\perp}^2\otimes \bP_{\bX^\perp}^3, \cr
\calQ_{\bX}^2 =  \bP_{\bX}^1\otimes \bP_{\bX^\perp}^2\otimes \bP_{\bX}^3, &&\calQ_{\bX^\perp}^2 =  \bP_{\bX^\perp}^1\otimes \bP_{\bX}^2\otimes \bP_{\bX^\perp}^3, \\ \nonumber
\calQ_{\bX}^3 =  \bP_{\bX}^1\otimes \bP_{\bX}^2\otimes \bP_{\bX^\perp}^3, && \calQ_{\bX^\perp}^3 =  \bP_{\bX^\perp}^1\otimes \bP_{\bX^\perp}^2\otimes \bP_{\bX}^3, \\ \nonumber
\calQ_{\bX} = \calQ_{\bX}^0+\calQ_{\bX}^1+\calQ_{\bX}^2+\calQ_{\bX}^3, && \calQ_{\bX^\perp} = \calQ_{\bX^\perp}^0+\calQ_{\bX^\perp}^1+\calQ_{\bX^\perp}^2+\calQ_{\bX^\perp}^3. 
\ees

\subsection{Subdifferential of Tensor Nuclear Norm}
One of the main technical tools in analyzing the nuclear norm minimization is the characterization of the subdifferntial of the nuclear norm. Such results are well known in the case of matrices. In particular, let $M=UDV^\top$ be the singular value decomposition of a matrix $M$, then the subdifferential of the nuclear norm at $M$ is given by
$$
\partial \|\cdot\|_\ast (M)=\{UV^\top+W: U^\top W=W^\top V=0,\quad {\rm and}\quad \|W\|\le1\},
$$
where with slight abuse of notion, $\|\cdot\|_\ast$ and $\|\cdot\|$ are the nuclear and spectral norms of matrices. 
%%%
See, e.g., Watson (1992). 
In other words, for any other matrix $Y$ of conformable dimensions,
$$
\|Y\|_\ast\ge \|M\|_\ast+\langle UV^\top+W, Y-M\rangle
$$
if and only if $U^\top W=W^\top V=0$ and $\|W\|\le1$. Characterizing the subdifferential of the nuclear norm for higher order tensors is more subtle due to the lack of corresponding spectral decomposition. 

A straightforward generalization of the above characterization may suggest that $\partial\|\cdot\|_\ast(\bX)$ be identified with
$$
\{ \bW+\bW^\perp: \bW^\perp =\calQ_{\bX^\perp}\bW^\perp, \|\bW^\perp\|\le 1\},
$$
for some $\bW$ in the range of $\calQ_{\bX}^0$. It turns out that this in general is not true. As a simple counterexample, let
$$\bX=\bfe_1\otimes \bfe_1\otimes \bfe_1,$$
and
$$\bY=\sum_{1\le i,j,k\le 2} \bfe_i\otimes \bfe_j\otimes \bfe_k=(\bfe_1+\bfe_2)\otimes (\bfe_1+\bfe_2)\otimes (\bfe_1+\bfe_2),$$
where $d_1=d_2=d_3=2$ and 
%%%
$\bfe_i$'s are the canonical basis of an Euclidean space. It is clear that $\|\bX\|_\ast=1$ and $\|\bY\|_\ast=2\sqrt{2}$. 
%%%
Take $\bW^\perp=\bU/\|\bU\|$ where
\bel{eq:defU}
\bU=\bfe_1\otimes \bfe_2\otimes \bfe_2+\bfe_2\otimes \bfe_1\otimes \bfe_2+\bfe_2\otimes \bfe_2\otimes \bfe_1.
\eel
%%%
As we shall show in the proof of Lemma \ref{pr:sub diff} below, $\|\bU\| = 2/\sqrt{3}$. 
It is clear that $\bW^\perp=\calQ_{\bX^\perp}\bW^\perp$ and $\|\bW^\perp\|\le 1$. Yet,
%%%
$$
\|\bX\|_\ast+\langle \bY-\bX,\bW+\bW^\perp\rangle=1+\langle \bY-\bX,\bW^\perp\rangle 
=1+ 3\sqrt{3}/2 >2\sqrt{2}=\|\bY\|_\ast,
$$
for any $\bW$ such that $\bW=\calQ_{\bX}^0\bW$.

Fortunately, for our purposes, the following relaxed characterization is sufficient.

\begin{lemma}
\label{pr:sub diff}
For any third order tensor $\bX\in \R^{d_1\times d_2\times d_3}$,  there exists a $\bW\in \R^{d_1\times d_2\times d_3}$ such that $\bW=\calQ_{\bX}^0\bW$, $\|\bW\|=1$ and
$$
\|\bX\|_\ast=\langle \bW, \bX\rangle.
$$
Furthermore, for any $\bY\in \R^{d_1\times d_2\times d_3}$ and $\bW^{\perp}\in \R^{d_1\times d_2\times d_3}$ obeying 
%%% $\bW^{\perp}=\calQ_{\bX^\perp}\bW^{\perp}$ and 
$\|\bW^{\perp}\|\le 1/2$,
$$
\|\bY\|_\ast\ge \|\bX\|_\ast+\langle \bW+\calQ_{\bX^\perp}\bW^\perp, \bY-\bX\rangle.
$$
\end{lemma}
\vskip 20pt

\noindent {\sc Proof of Lemma \ref{pr:sub diff}.}
If $\|\bW\|\le 1$, then 
$$
\max_{\|\bu_j\|=1}\langle \calQ_{\bX}^0\bW,\bu_1\otimes\bu_2\otimes \bu_3\rangle\le \max_{\|\bu_j\|=1}\langle \bW, (\bP_{\bX}^1\bu_1)\otimes(\bP_{\bX}^2\bu_2)\otimes (\bP_{\bX}^3\bu_3)\rangle \le 1. 
$$
It follows that
$$\|\bX\|_*=\max_{\|\bW\|=1}\langle \bW,\bX\rangle=\max_{\|\bW\|=1}\langle \calQ_{\bX}^0\bW,\bX\rangle$$
is attained with a certain $\bW$ satisfying $\|\bW\|=\|\calQ_{\bX}^0\bW\|=1$. 

Now consider a tensor $\bW^{\perp}$ satisfying
$$\|\bW+\calQ_{\bX^\perp}\bW^{\perp}\|\le 1.$$ 
Because $\calQ_{\bX^\perp}\bX=0$, it follows from the definition of the tensor nuclear norm that
$$
\langle \bW+\calQ_{\bX^\perp}\bW^\perp, \bY-\bX\rangle
\le \|\bY\|_* - \langle \bW, \bX\rangle = \|\bY\|_\ast - \|\bX\|_\ast. 
$$
It remains to prove that $\|\bW^{\perp}\|\le 1/2$ implies 
$$\|\bW+\calQ_{\bX^\perp}\bW^{\perp}\|\le 1.$$ 
Recall that $\bW=\calQ_{\bX}^0\bW$, $\|\bW^{\perp}\|\le 1/2$, and $\|\bu_j\|=1$. Then
\bes
&& \langle \bW+\calQ_{\bX^\perp}\bW^{\perp},\bu_1\otimes\bu_2\otimes \bu_3\rangle\cr
&\le& \|\calQ_{\bX}^0(\bu_1\otimes\bu_2\otimes \bu_3)\|_*+{1\over 2}\|\calQ_{\bX^\perp}(\bu_1\otimes\bu_2\otimes \bu_3)\|_*\cr 
&\le& \prod_{j=1}^3\sqrt{1-a_j^2} + {1\over 2}\Big(a_1a_2+a_1a_3\sqrt{1-a_2^2}+a_2a_3\sqrt{1-a_1^2}\Big). 
\ees
where $a_j = \|\bP_{\bX^\perp}^j\bu_j\|_2$, for $j=1,2,3$. Let $x=a_1a_2$ and
$$y=\sqrt{(1-a_1^2)(1-a_2^2)}.$$
We have 
\bes
\Big(a_1\sqrt{1-a_2^2}+a_2\sqrt{1-a_1^2}\Big)^2
&=& a_1^2(1-a_2^2)+a_2^2(1-a_1^2)+2xy
\cr &=& a_1^2+a_2^2 - 2a_1^2a_2^2 + 2xy
\cr &=& 1 - (y-x)^2.
\ees
It follows that for any value of $a_3\in (0,1)$, 
$$
\langle \bW+\calQ_{\bX^\perp}\bW^{\perp},\bu_1\otimes\bu_2\otimes \bu_3\rangle\le y \sqrt{1-a_3^2}+{1\over 2}\Big(x+a_3\sqrt{1-(y-x)^2}\Big). 
$$
This function of $(x,y)$ is increasing in the smaller of $x$ and $y$. For $x<y$, the maximum of $x^2$ given $y^2$ is attained when $a_1=a_2$ by simple calculation with the Lagrange multiplier. Similarly, for $y<x$, the maximum of $y^2$ given $x^2$ is attained when $a_1=a_2$. Thus, setting $a_1=a_2=a$, we find 
\bes
\langle \bW+\calQ_{\bX^\perp}\bW^{\perp},\bu_1\otimes\bu_2\otimes \bu_3\rangle
\le \max_{a_3,a}\Big\{(1-a^2)\sqrt{1-a_3^2}+{1\over 2}\Big(a^2+2a_3 a\sqrt{1 - a^2}\Big)\Big\}. 
\ees
The above maximum is attained when $a_3=a$. Because $\sqrt{1-a^2} + a^2/2\le 1$, we have 
$$\langle \bW+\calQ_{\bX^\perp}\bW^{\perp},\bu_1\otimes\bu_2\otimes \bu_3\rangle\le 1,$$ 
%%%
which completes the proof of the lemma. 

%%%
The norm of $\bU$ defined in (\ref{eq:defU}) can be computed using a similar argument:
\bes
\|\bU\| 
&=& \max_{0\le a_1\le a_2\le a_3\le 1}\Big(a_1a_2\sqrt{1-a_3^2}+a_1a_3\sqrt{1-a_2^2}+a_2a_3\sqrt{1-a_1^2}\Big)
\cr &=& \max_{x,y,a_3}\Big(x\sqrt{1-a_3^2}+a_3\sqrt{1-(y-x)^2}\Big)
\cr &=& \max_{a,a_3}a\Big(a\sqrt{1-a_3^2}+2a_3\sqrt{1-a^2}\Big),
\cr &=& \max_{a}a\sqrt{a^2+4(1-a^2)},
\ees
which yields $\|\bU\|=2/\sqrt{3}$ with $a^2 = 2/3$. 
$\hfill\square$

\vskip 20pt 

Note that Lemma \ref{pr:sub diff} gives only sufficient conditions of the subgradient of tensor nuclear norm. Equivalently it states that
$$
%%% 
\partial\|\cdot\|_\ast(\bX)\supseteq\big\{ \bW+\calQ_{\bX^\perp}\bW^\perp: \|\bW^\perp\|\le 1/2\big\}. 
$$
%%%
The main difference between the above characterization and that for the usual matrix nuclear norm is the requirement that $\|\bW^\perp\|$ is no greater than $1/2$ instead of $1$. We note also that the constant $1/2$ may be further improved. No attempt has been made here to sharpen the constant as it already suffices for our analysis.

%%%\subsection{Incoherence}
\subsection{Coherence}

A central concept to matrix completion is coherence. 
Recall that the coherence of an $r$ dimensional linear subspace $U$ of $\R^k$ is defined as
%%%
$$
\mu(U)={k\over r}\max_{1\le i\le k}\|\bP_U\bfe_i\|^2 
= \frac{\max_{1\le i\le k}\|\bP_U\bfe_i\|^2}{k^{-1}\sum_{i=1}^k\|\bP_U\bfe_i\|^2}, 
$$
where $\bP_U$ is the orthogonal projection onto $U$ and 
%%%
$\bfe_i$'s are the canonical basis for $\R^k$. See, e.g., Cand\`es and Recht (2008). We shall define the coherence of a tensor $\bX\in \R^{d_1\times d_2\times d_3}$ as
$$\mu(\bX)=\max\{\mu(\calL_1(\bX)),\mu(\calL_2(\bX)),\mu(\calL_3(\bX))\}.$$%\max\{\mu(\calL_1(\bX))\mu(\calL_2(\bX)),\mu(\calL_2(\bX))\mu(\calL_3(\bX)),\mu(\calL_1(\bX))\mu(\calL_3(\bX))\}.$$
%%%
It is clear that $\mu(\bX)\ge 1$, 
since $\mu(U)$ is the ratio of the $\ell_\infty$ and length-normalized $\ell_2$ norms of a vector. 

\begin{lemma}\label{lm-mu}
Let $\bX\in \R^{d_1\times d_2\times d_3}$ be a third order tensor. Then
$$
\max_{a,b,c}\|\calQ_{\bX}(\bfe_a\otimes\bfe_b\otimes\bfe_c)\|_{\rm HS}^2\le {\rbar^2(\bX)d\over d_1d_2d_3}\mu^2(\bX).
$$
\end{lemma}
\vskip 20pt

\noindent{\sc Proof of Lemma \ref{lm-mu}.}
Recall that $\calQ_{\bX}=\calQ_{\bX}^0+\calQ_{\bX}^1+\calQ_{\bX}^2+\calQ_{\bX}^3$. Therefore,
\bes
\|\calQ_{\bX}(\bfe_a\otimes \bfe_b\otimes \bfe_c)\|^2&=&\sum_{j,k=0}^3\langle \calQ_{\bX}^j(\bfe_a\otimes \bfe_b\otimes \bfe_c),\calQ_{\bX}^k(\bfe_a\otimes \bfe_b\otimes \bfe_c)\rangle\\
&=&\sum_{j=0}^3\|\calQ_{\bX}^j(\bfe_a\otimes \bfe_b\otimes \bfe_c)\|^2\\
&=&\|\bP_{\bX}^1\bfe_a\|^2\|\bP_{\bX}^2\bfe_b\|^2\|\bP_{\bX}^3\bfe_c\|^2+\|\bP_{\bX^\perp}^1\bfe_a\|^2\|\bP_{\bX}^2\bfe_b\|^2\|\bP_{\bX}^3\bfe_c\|^2\\
&&+\|\bP_{\bX}^1\bfe_a\|^2\|\bP_{\bX^\perp}^2\bfe_b\|^2\|\bP_{\bX}^3\bfe_c\|^2+\|\bP_{\bX}^1\bfe_a\|^2\|\bP_{\bX}^2\bfe_b\|^2\|\bP_{\bX^\perp}^3\bfe_c\|^2.
\ees
For brevity, write $r_j=r_j(\bX)$, and $\mu=\mu(\bX)$. Then
\bes
\|\bP_{\bX}^1\bfe_a\|^2\|\bP_{\bX}^2\bfe_b\|^2\|\bP_{\bX}^3\bfe_c\|^2+\|\bP_{\bX^\perp}^1\bfe_a\|^2\|\bP_{\bX}^2\bfe_b\|^2\|\bP_{\bX}^3\bfe_c\|^2&\le& {r_2r_3\mu^2\over d_2d_3};\\
\|\bP_{\bX}^1\bfe_a\|^2\|\bP_{\bX}^2\bfe_b\|^2\|\bP_{\bX}^3\bfe_c\|^2+\|\bP_{\bX}^1\bfe_a\|^2\|\bP_{\bX^\perp}^2\bfe_b\|^2\|\bP_{\bX}^3\bfe_c\|^2&\le& {r_1r_3\mu^2\over d_1d_3};\\
\|\bP_{\bX}^1\bfe_a\|^2\|\bP_{\bX}^2\bfe_b\|^2\|\bP_{\bX}^3\bfe_c\|^2+\|\bP_{\bX}^1\bfe_a\|^2\|\bP_{\bX}^2\bfe_b\|^2\|\bP_{\bX^\perp}^3\bfe_c\|^2&\le& {r_1r_2\mu^2\over d_1d_2}.
\ees
As a result, for any $(a,b,c)\in [d_1]\times[d_2]\times[d_3]$,
$$
\|\calQ_{\bX}(\bfe_a\otimes \bfe_b\otimes \bfe_c)\|^2\le {\mu^2(r_1r_2d_3+r_1d_2r_3+d_1r_2r_3)\over d_1d_2d_3},
$$
which implies the desired statement. $\hfill\square$
\vskip 20pt

Another measure of coherence for a tensor $\bX$ is
$$\alpha(\bX):=\sqrt{d_1d_2d_3/\rbar(\bX)}\|\bW\|_{\max}$$
where $\bW$ is such that $\bW=\calQ_{\bX}^0\bW$, $\|\bW\|=1$ and $\langle \bX, \bW\rangle =\|\bX\|_\ast$ 
as described in Lemma~\ref{pr:sub diff}. 
%%%
The quantity $\alpha(\bX)$ is related to $\mu(\bX)$ defined earlier and 
the spikiness 
\bes
\talpha(\bX) = \sqrt{d_1d_2d_3}\|\bW\|_{\max}/\|\bW\|_{\rm HS}. 
\ees

\begin{lemma}\label{lm-alpha}
Let $\bX\in \R^{d_1\times d_2\times d_3}$ be a third order tensor. 
Assume without loss of generality that $r_1(\bX)\le r_2(\bX)\le r_3(\bX)$. 
Then,  
%%%
$$
\alpha^2(\bX)\le \min\Big\{r_1(\bX)r_2(\bX)r_3(\bX)\mu^{3}(\bX)/\rbar(\bX),r_1(\bX)r_2(\bX)\talpha^2(\bX)/\rbar(\bX)\Big\}.
$$
Moreover, if $\bX$ admits a bi-orthogonal eigentensor decomposition 
$\sum_{i=1}^r \lam_i (\bu_i\otimes \bv_i\otimes \bw_i)$ 
%%%%
with $\lam_i\neq 0$ and 
$\bu_i^\top\bu_j=\bv_i^\top\bv_j\bw_i^\top\bw_j={\mathbb I}\{i=j\}$ for $1\le i,j\le r$, 
then $r_1(\bX) = r_3(\bX) = \rbar(\bX) = r$, $\|X^{(1)}\|_*=\|\bX\|_*$, and 
$$
\alpha(\bX)=\talpha(\bX)\ge 1. 
$$
\end{lemma}
\vskip 20pt

%%%
%The above upper bound of $\alpha(\bX)$ in terms of $\talpha(\bX)$ is equivalent to 
%$$\|\bW\|_{\rm HS}\le \sqrt{r_1(\bX)r_2(\bX)}.$$

\noindent{\sc Proof of Lemma \ref{lm-alpha}.}
%%% 
Due to the conditions $\bW=\calQ_{\bX}^0\bW$ and $\|\bW\|=1$, 
\bes
\|\bW\|_{\max}^2
&=& \max_{a,b,c}|\langle \bW,(\bP_{\bX}^1\bfe_a)\otimes(\bP_{\bX}^2\bfe_b)\otimes(\bP_{\bX}^3\bfe_c)\rangle|^2
\cr &\le& \max_{a,b,c}\|\bP_{\bX}^1\bfe_a\|^2\|\bP_{\bX}^2\bfe_b\|^2\|\bP_{\bX}^3\bfe_c\|^2
\cr &\le & r_1(\bX)r_2(\bX)r_3(\bX)\mu^{3}(\bX)/(d_1d_2d_3),
\ees
which yields the upper bound for $\alpha(\bX)$ in terms of $\mu(\bX)$. 

Because $\bW$ is in the range of $\calQ_{\bX}^0$, $\calL_j(\bW)\subseteq \calL_j(\bX)$. Therefore, $r_1(\bW)\le r_1(\bX)$. Recall that $W^{(1)}$ is a $d_1\times (d_2d_3)$ matrix whose columns are the mode-1 fibers of $\bW$. Applying singular value decomposition to $W^{(1)}$ suggests that there are orthornomal vectors $\{\bu_1,\ldots,\bu_{r_1}\}$ in $\R^{d_1}$ and matrices $M_1,\ldots, M_{r_1}\in \R^{d_2\times d_3}$ such that $\langle M_j, M_k\rangle=0$ if $j\neq k$, and
$$
\bW=\sum_{k=1}^{r_1(\bX)}\bu_k\otimes M_k.
$$
It is clear that $\|M_k\|\le \|\bW\|=1$, and $\rank(M_k)\le r_2(\bX)$. Therefore, 
$$
\|\bW\|_{\rm HS}^2 
\le \sum_{k=1}^{r_1(\bX)}\|M_k\|_{\rm HS}^2 \le r_1(\bX)r_2(\bX). 
$$
This gives the upper bound for $\alpha(\bX)$ in terms of $\talpha(\bX)$. 

It remains to consider the case of $\bX = \sum_{i=1}^r \lam_i (\bu_i\otimes \bv_i\otimes \bw_i)$. 
%%%%
%Assume $\lam_i > 0$ without loss of generality. 
Obviously, by triangular inequality,
$$
\|\bX\|_\ast\le \sum_{i=1}^r \|\bu_i\otimes \bv_i\otimes \bw_i\|_\ast=\sum_{i=1}^r \|\bw_i\|.
$$
On the other hand, let
$$\bW = \sum_{i=1}^r \bu_i\otimes \bv_i\otimes (\bw_i/\|\bw_i\|).$$ 
Because
$$
\|\bW\| \le \max_{\ba,\bb,\bc: \|\ba\|,\|\bb\|,\|\bc\|\le 1} \sum_{i=1}^n \Big|\big(\ba^\top\bu_i\big) \trace(\bc \bb^\top\bv_i\bw_i^\top/\|\bw_i\|)\Big| 
\le \|\ba\| \|\bb\bc^\top\|_{\rm HS}\le 1,
$$
we find
$$
\|\bX\|_\ast\ge \langle \bW,\bX\rangle=\sum_{i=1}^r \|\bw_i\|,
$$
which implies that $\bW$ is dual to $\bX$ and
$$
\|\bX\|_\ast=\sum_{i=1}^r \|\bw_i\|,
$$
where the rightmost hand side also equals to $\|X^{(1)}\|_\ast$ and $\|X^{(2)}\|_\ast$. The last statement now follows from the fact that $\|\bW\|_{\rm HS}^2=r$.
$\hfill\square$
\vskip 20pt

As in the matrix case, exact recovery with observations on a small fraction of the entries is only possible for 
%%%
tensors with low coherence. In particular, we consider in this article the recovery of a tensor $\bT$ 
obeying $\mu(\bT)\le \mu_0$ and $\alpha(\bT)\le \alpha_0$ for some $\mu_0,\alpha_0\ge 1$.

\section{Exact Tensor Recovery}
\label{sec:main}
We are now in position to study the nuclear norm minimization for tensor completion. Let $\hbT$ be the solution to
\bel{eq:defnnm}
\min_{\bX\in \R^{d_1\times d_2\times d_3}} \|\bX\|_\ast\qquad {\rm subject\ to\ } \calP_\Omega \bX=\calP_\Omega \bT,
\eel
where $\calP_\Omega: \R^{d_1\times d_2\times d_3}\mapsto \R^{d_1\times d_2\times d_3}$ such that
$$
(\calP_\Omega \bX)(i,j,k)=\left\{\begin{array}{ll}\bX(i,j,k)& {\rm if\ }(i,j,k)\in\Omega\\ 0& {\rm otherwise}\end{array}\right..
$$
%%%
Assume that $\Omega$ is a uniformly sampled subset of $[d_1]\times [d_2]\times [d_3]$. 
The goal is to determine what the necessary sample size is for successful reconstruction of $\bT$ using $\hbT$ with high probability. In particular, we show that that with high probability, exact recovery can be achieved with nuclear norm minimization (\ref{eq:defnnm}) if
$$
|\Omega|\ge \left(\alpha_0\sqrt{rd_1d_2d_3}+\mu^2_0 r^2 d\right)\polylog(d),
$$
where $d=d_1+d_2+d_3$. More specifically, we have 

%%%%
\begin{theorem}\label{th-1}
Assume that $\mu(\bT)\le \mu_0$, $\alpha(\bT)\le \alpha_0$, and $\rbar(\bT)=r$. 
Let $\Omega$ be a uniformly sampled subset of $[d_1]\times[d_2]\times[d_3]$ and 
$\hbT$ be the solution to (\ref{eq:defnnm}). For $\beta>0$, define 
\bes
q^*_1=  \big(\beta + \log d\big)^2\alpha_0^2r\log d,\quad q^*_2=(1+\beta)(\log d)\mu_0^2r^2. 
\ees
Let $n=|\Omega|$. Suppose that for a sufficiently large numerical constant $c_0$, 
\bel{case-1}
n \ge c_0\delta_2^{-1}\left[\sqrt{q^*_1(1+\beta)\delta_1^{-1}d_1d_2d_3}+q^*_1d^{1+\delta_1}
+ q^*_2d^{1+\delta_2}\right]
\eel
with certain $\{\delta_1,\delta_2\}\in [1/\log d,1/2]$ and $\beta>0$. Then, 
\bel{eq:rec}
\P\Big\{ \hbT \neq \bT\Big\}\le d^{-\beta}.
\eel
In particular, for $\delta_1=\delta_2=(\log d)^{-1}$, 
(\ref{case-1}) can be written as 
\bes
n\ge C_{\mu_0,\alpha_0,\beta}\Big[(\log d)^3\sqrt{rd_1d_2d_3}+\big\{r(\log d)^3+r^2(\log d)\big\}d\Big]
\ees
with a constant $C_{\mu_0,\alpha_0,\beta}$ depending on $\{\mu_0,\alpha_0,\beta\}$ only. 
\end{theorem}

For $d_1\asymp d_2\asymp d_3$ and fixed $\{\alpha_0,\mu_0,\delta_1,\delta_2,\beta\}$, 
the sample size requirement (\ref{case-1}) becomes
\bes
n \asymp \sqrt{r}(d\log d)^{3/2}, 
\ees
provided $\max\{r(\log d)^3d^{2\delta_1},r^3d^{2\delta_2}/(\log d)\}=O(d)$. 

The high level idea of our strategy is similar to the matrix case -- exact recovery of $\bT$ is implied by the existence of 
%%% 
a dual certificate $\bG$ supported on $\Omega$, that is $\calP_\Omega \bG=\bG$, such that $\calQ_{\bT} \bG=\bW$ and $\|\calQ_{\bT^\perp}\bG\|<1/2$.

\subsection{Recovery with a Dual Certificate}
%%%
Write $\hbT=\bT+\Delta$. Then, $\calP_\Omega \Delta=0$ and
$$
\|\bT+\Delta\|_\ast\le \|\bT\|_\ast.
$$
Recall that, by Lemma \ref{pr:sub diff}, there exists a $\bW$ obeying $\bW=\calQ_{\bT}^0\bW$ and $\|\bW\|=1$ such that
%%%
$$
\|\bT+\Delta\|_\ast\ge \|\bT\|_\ast+\langle\bW+\calQ_{\bT^{\perp}}\bW^{\perp},\Delta\rangle
$$
for any $\bW^{\perp}$ obeying %$\bW^{\perp}=\calQ_{\bT^{\perp}}\bW^{\perp}$ and 
$\|\bW^{\perp}\|\le 1/2$. Assume that a tensor $\bG$ supported on $\Omega$, that is $\calP_\Omega \bG=\bG$, such that $\calQ_{\bT} \bG=\bW$, and $\|\calQ_{\bT^\perp}\bG\|<1/2$. 
%%%
When $\|\calQ_{\bT^\perp}\Delta\|_\ast>0$, 
\begin{eqnarray*}
\langle\bW+\calQ_{\bT^{\perp}}\bW^{\perp},\Delta\rangle&=&\langle\bW+\calQ_{\bT^{\perp}}\bW^{\perp}-\bG,\Delta\rangle\\
&=&\langle \bW-\calQ_{\bT}\bG, \Delta\rangle+\langle \bW^\perp, \calQ_{\bT^\perp}\Delta\rangle-\langle \calQ_{\bT^\perp}\bG, \calQ_{\bT^\perp}\Delta\rangle\\
&>&\langle \bW^\perp, \calQ_{\bT^\perp}\Delta\rangle-{1\over 2}\|\calQ_{\bT^\perp}\Delta\|_\ast
\end{eqnarray*}
%%%
Take $\bW^\perp=\bU/2$ where
$$\bU=\argmax_{\bX: \|\bX\|\le 1}\langle \bX, \calQ_{\bT^\perp}\Delta\rangle.$$
%%%
We find that $\|\calQ_{\bT^\perp}\Delta\|_\ast>0$ implies 
%%%
$$
\|\bT+\Delta\|_\ast - \|\bT\|_\ast \ge \langle\bW+\calQ_{\bT^\perp}\bW^{\perp},\Delta\rangle> 0,
$$
which contradicts with fact that $\hbT$ minimizes the nuclear norm. 
%%%
Thus, $\calQ_{\bT^\perp}\Delta = 0$, which then implies 
$\calQ_{\bT}\calQ_{\Omega}\calQ_{\bT}\Delta = \calQ_{\bT}\calQ_{\Omega}\Delta=0$. 
When $\calQ_{\bT}\calQ_{\Omega}\calQ_{\bT}$ is invertible in the range of $\calQ_{\bT}$, 
we also have $\calQ_{\bT}\Delta=0$ and $\hbT=\bT$.

With this in mind, it then suffices to seek such a dual certificate. 
In fact, it turns out that finding an ``approximate'' dual certificate is actually enough for our purposes. 

\begin{lemma}\label{prop-dual-cert} 
Assume that
\bel{eq:proj-on-T}
\inf\Big\{\|\calP_\Omega\calQ_{\bT}\bX\|_{\rm HS}: \|\calQ_{\bT}\bX\|_{\rm HS}=1\Big\}\ge \sqrt{n\over 2d_1d_2d_3}. 
\eel
If there exists a tensor $\bG$ supported on $\Omega$ such that
\bel{dual-cert}
%%%%
\|\calQ_{\bT}\bG-\bW\|_{\rm HS} < \frac{1}{4}\sqrt{n\over 2d_1d_2d_3}\qquad {\rm and}\qquad 
\max_{\|\calQ_{\bT^\perp}\bX\|_*=1}\langle\bG,\calQ_{\bT^{\perp}}\bX\rangle \le 1/4,
\eel
then $\hbT=\bT$.  
\end{lemma} 
\vskip 20pt

\noindent{\sc Proof of Lemma \ref{prop-dual-cert}.}
Write $\hbT=\bT+\bDelta$, then $\calP_\Omega \bDelta=0$ and
$$
\|\bT+\bDelta\|_\ast\le \|\bT\|_\ast.
$$
Recall that, by Lemma \ref{pr:sub diff}, there exists a $\bW$ obeying $\bW=\calQ_{\bT}^0\bW$ and $\|\bW\|=1$ such that for any $\bW^{\perp}$ obeying $\|\bW^{\perp}\|\le 1/2$, 
$$
\|\bT+\bDelta\|_\ast\ge \|\bT\|_\ast+\langle\bW+\calQ_{\bT^{\perp}}\bW^{\perp},\bDelta\rangle. 
$$
Since $\langle\bG,\bDelta\rangle = \langle\calP_\Omega\bG,\bDelta\rangle 
= \langle\bG,\calP_\Omega\bDelta\rangle=0$ and $\calQ_{\bT}\bW=\bW$, 
\bes
0 &\ge& \langle\bW+\calQ_{\bT^{\perp}}\bW^{\perp},\bDelta\rangle
\cr &=&%\langle\bG,\bDelta\rangle+
\langle\bW+\calQ_{\bT^{\perp}}\bW^{\perp}-\bG,\bDelta\rangle\\
&=&\langle \calQ_{\bT}\bW-\calQ_{\bT}\bG, \bDelta\rangle
+\langle \bW^{\perp}, \calQ_{\bT^{\perp}}\bDelta\rangle-\langle\bG, \calQ_{\bT^{\perp}}\bDelta\rangle\\
&\ge&-\|\bW-\calQ_{\bT}\bG\|_{\rm HS}\|\calQ_{\bT}\bDelta\|_{\rm HS}
+\langle \bW^{\perp}, \calQ_{\bT^{\perp}}\bDelta\rangle-{1\over 4}\|\calQ_{\bT^{\perp}}\bDelta\|_\ast.
\ees
In particular, taking $\bW^{\perp}$ satisfying $\|\bW\|=1/2$ and 
$\langle \bW^{\perp}, \calQ_{\bT^{\perp}}\bDelta\rangle
= \|\calQ_{\bT^{\perp}}\bDelta\|_\ast/2$, we find 
\bes
{1\over 4}\|\calQ_{\bT^{\perp}}\bDelta\|_\ast
\le \|\bW-\calQ_{\bT}\bG\|_{\rm HS}\|\calQ_{\bT}\bDelta\|_{\rm HS}.
\ees

Recall that
$\calP_\Omega\bDelta=\calP_\Omega\calQ_{\bT^{\perp}}\bDelta+\calP_\Omega\calQ_{\bT}\bDelta=0$. 
Thus, in view of the condition on $\calP_\Omega$, % and Lemma \ref{prop-1}
\bel{eq:hsbd}
 \frac{\|\calQ_{\bT}\bDelta\|_{\rm HS}}{\sqrt{2d_1d_2d_3/n}}
 \le \|\calP_\Omega\calQ_{\bT}\bDelta\|_{\rm HS}=\|\calP_\Omega\calQ_{\bT^{\perp}}\bDelta\|_{\rm HS}
 \le \|\calQ_{\bT^{\perp}}\bDelta\|_{\rm HS}
 \le \|\calQ_{\bT^{\perp}}\bDelta\|_*. 
\eel
Consequently, 
 \bes
{1\over 4}\|\calQ_{\bT^{\perp}}\bDelta\|_\ast
\le \sqrt{2d_1d_2d_3/n}\|\bW-\calQ_{\bT}\bG\|_{\rm HS}\|\calQ_{\bT^{\perp}}\bDelta\|_\ast.
\ees
Since
$$\sqrt{2d_1d_2d_3/n}\|\bW-\calQ_{\bT}\bG\|_{\rm HS}<1/4,$$ 
we have $\|\calQ_{\bT^{\perp}}\bDelta\|_\ast=0$. 
Together with (\ref{eq:hsbd}), we conclude that $\bDelta=0$, or equivalently $\hbT=\bT$. $\hfill\square$
\vskip 20pt

Equation (\ref{eq:proj-on-T}) indicates the invertibility of $\calP_\Omega$ when restricted to the range of $\calQ_{\bT}$. We argue first that this is true for ``incoherent'' tensors. To this end, we prove that
\bes
\Big\|\calQ_{\bT}\Big((d_1d_2d_3/n)\calP_\Omega - \calI\Big)\calQ_{\bT}\Big\| \le 1/2
\ees
with high probability. This implies that as an operator in the range of $\calQ_{\bT}$, the spectral norm of $(d_1d_2d_3/n)\calQ_{\bT}\calP_\Omega\calQ_{\bT}$ is contained in $[1/2,3/2]$. Consequently, (\ref{eq:proj-on-T}) holds because for any $\bX\in \R^{d_1\times d_2\times d_3}$,
\bes
(d_1d_2d_3/n)\|\calP_\Omega\calQ_{\bT}\bX\|_{\rm HS}^2 =  
\Big\langle \calQ_{\bT}\bX, (d_1d_2d_3/n)\calQ_{\bT}\calP_\Omega\calQ_{\bT}\bX\Big\rangle 
\ge \frac{1}{2}\|\calQ_{\bT}\bX\|_{\rm HS}^2.
\ees
Recall that $d=d_1+d_2+d_3$. We have

\begin{lemma}\label{lm:op-norm}
Assume $\mu(\bT)\le \mu_0$, $\rbar(\bT)=r$, and $\Omega$ is uniformly sampled from $[d_1]\times[d_2]\times [d_3]$ without replacement. Then, for any $\tau>0$,
$$
\P\left\{\left\|\calQ_{\bT}\big((d_1d_2d_3/n)\calP_\Omega - \calI\big)\calQ_{\bT}\right\|\ge \tau\right\}
\le 2r^2 d\,\exp\Big(-\frac{ \tau^2/2}{1+2\tau /3}\Big(\frac{n}{\mu^2_0 r^2 d}\Big)\Big).
$$
\end{lemma}

In particular, taking $\tau=1/2$ in Lemma \ref{lm:op-norm} yields
%$$
%\left\|\calQ_{\bT}\big((d_1d_2d_3/n)\calP_\Omega - \calI\big)\calQ_{\bT}\right\|\le 1/2
%$$
%with probability at least
%$$
%1-2r^2 d\,\exp\Big(-\frac{3}{32}\Big(\frac{n}{\mu^2_0 r^2 d}\Big)\Big).
%$$
\bes
\P\Big\{\hbox{ (\ref{eq:proj-on-T}) holds }\Big\} 
\ge 1-2r^2 d\,\exp\Big(-\frac{3}{32}\Big(\frac{n}{\mu^2_0 r^2 d}\Big)\Big).
\ees
\subsection{Constructing a Dual Certificate}
We now show that the ``approximate'' dual certificate as required by Lemma \ref{prop-dual-cert} can indeed be constructed when $\Omega$ is a uniformly sampled subset of $[d_1]\times[d_2]\times[d_3]$. We use a strategy similar to the ``golfing scheme'' for the matrix case (see, e.g., Gross, 2011).

We begin by constructing an iid uniformly distributed sequence in $[d_1]\times[d_2]\times[d_3]$, $\{(a_i,b_i,c_i): 1\le i\le n\}$. This can be done by sampling with replacement from $\Omega$:
\vskip 10pt
%\newpage
\noindent\hrulefill

\centerline{Creating IID Samples from $\Omega$}

\noindent\hrulefill
%\hline
\begin{itemize}
\item Initialize $S_0=\emptyset$.
\item For each $i=1,2,\ldots, n$, 
\begin{itemize}
\item with probability $|S_{i-1}|/d_1d_2d_3$, sample $(a_i,b_i,c_i)$ uniformly from $S_{i-1}$; and with probability $1-|S_{i-1}|/d_1d_2d_3$, sample $(a_i,b_i,c_i)$ 
%%%
uniformly from $\Omega\setminus S_{i-1}$.
\item Update $S_i = S_{i-1}\cup\{(a_i,b_i,c_i)\}$
\end{itemize}
\end{itemize}
Because $\P\{(a_i,b_i,c_i)\in S_{i-1}|S_{i-1}\}$ matches that of the iid case and $(a_i,b_i,c_i)$ is uniform in $[d_1]\times[d_2]\times[d_3]\setminus S_{i-1}$ conditionally on $S_{i-1}$ and $(a_i,b_i,c_i)\not\in S_{i-1}$, the points $(a_i,b_i,c_i)$ are iid uniform in $[d_1]\times[d_2]\times[d_3]$. 

\noindent\hrulefill

\vskip 10pt

We now divide the sequence $\{(a_i,b_i,c_i): 1\le i\le n\}$ into $n_2$ subsequences of length $n_1$:
$$
\Omega_k=\left\{(a_i,b_i,c_i): (k-1)n_1<i\le kn_1\right\},
$$
for $k=1,2,\ldots, n_2$, where $n_1n_2\le n$. Recall that $\bW$ is such that $\bW=\calQ_{\bT}^0 \bW$, $\|\bW\|=1$, and $\|\bT\|_\ast=\langle \bT, \bW\rangle$. Let 
$$
\calR_k = \calI - \frac{1}{n_1}\sum_{i=(k-1)n_1+1}^{kn_1}(d_1d_2d_3)\calP_{(a_i,b_i,c_i)}
$$
with $\calI$ being the identity operator on tensors and define 
$$
\bG_k = \sum_{\ell=1}^k \big(\calI - \calR_\ell\big)\calQ_{\bT}\calR_{\ell-1}\calQ_{\bT}\cdots\calQ_{\bT}\calR_{1}\calQ_{\bT}\bW,\quad \bG = \bG_{n_2}.
$$
Since $(a_i,b_i,c_i)\in\Omega$, $\calP_\Omega(\calI- \calR_k)=\calI - \calR_k$, 
so that $\calP_\Omega\bG = \bG$. It follows from the definition of $\bG_k$ that 
\bes
\calQ_{\bT}\bG_k 
&=& \sum_{\ell=1}^{k}(\calQ_{\bT}- \calQ_{\bT}\calR_\ell\calQ_{\bT})
(\calQ_{\bT}\calR_{\ell-1}\calQ_{\bT})\cdots(\calQ_{\bT}\calR_{1}\calQ_{\bT}\bW)
\cr &=& \bW - (\calQ_{\bT}\calR_{k}\calQ_{\bT})\cdots(\calQ_{\bT}\calR_1\calQ_{\bT})\bW
\ees
and 
\bes
\langle\bG_k,\calQ_{\bT^{\perp}}\bX\rangle 
= \Big\langle \sum_{\ell=1}^k \calR_\ell(\calQ_{\bT}\calR_{\ell-1}\calQ_{\bT})
\cdots(\calQ_{\bT}\calR_{1}\calQ_{\bT})\bW,\calQ_{\bT^{\perp}}\bX\Big\rangle. 
\ees
Thus, condition (\ref{dual-cert}) holds if 
\bel{dual-cert-1}
\|(\calQ_{\bT}\calR_{n_2}\calQ_{\bT})\cdots(\calQ_{\bT}\calR_1\calQ_{\bT})\bW\|_{\rm HS} 
< \frac{1}{4}\sqrt{n\over 2d_1d_2d_3}
\eel
and 
\bel{dual-cert-2}
\Big\|\sum_{\ell=1}^{n_2} \calR_\ell(\calQ_{\bT}\calR_{\ell-1}\calQ_{\bT})
\cdots(\calQ_{\bT}\calR_{1}\calQ_{\bT})\bW\Big\| <1/4. 
\eel

\subsection{Verifying Conditions for Dual Certificate}

We now prove that (\ref{dual-cert-1}) and (\ref{dual-cert-2}) hold with high probability for the approximate dual certificate constructed above. For this purpose, we need large deviation bounds for the average of certain iid tensors under the spectral and maximum norms.

\begin{lemma}\label{lm-tensor-ineq}
Let $\{(a_i,b_i,c_i)\}$ be an independently and uniformly sampled sequence from $[d_1]\times [d_2]\times[d_3]$. Assume that $\mu(\bT)\le \mu_0$ and $\rbar(\bT)=r$. Then, for any fixed $k=1,2,\ldots,n_2$, and for all $\tau>0$, 
\bel{tensor-ineq-1} 
\P\Big\{\Big\|\calQ_{\bT}\calR_k\calQ_{\bT}\Big\| \ge \tau \Big\}
\le 2r^2 d\exp\Big(-\frac{ \tau^2/2}{1+2\tau /3}\Big(\frac{n_1}{\mu^2_0 r^2 d}\Big)\Big),
\eel
and
\bel{tensor-ineq-2}
\max_{\|\bX\|_{\max}=1}\P\Big\{\Big\|\calQ_{\bT}\calR_k\calQ_{\bT}\bX\Big\|_{\max} \ge \tau\Big\} 
\le 2d_1d_2d_3\,\exp\Big(-\frac{ \tau^2/2}{1+2\tau /3}\Big(\frac{n_1}{\mu^2_0 r^2 d}\Big)\Big). 
\eel
\end{lemma}

%\bes
%n &\ge& c_2(1+k_0)(1+\beta)\max\Big(\sqrt{rd_1d_2d_3}(\log d)^{3/2}, ((\mu_0r)\vee \sqrt{r_1r_2})^{1/k_0}\mu_0^2r^2d\log d\Big)
%\cr \sqrt{n} &\ge& \frac{d^3}{d_1d_2d_3(r\log d)^{3/2}}. 
%\ees
%We may also replace $1+k_0$ by $\log((\mu_0r)\vee \sqrt{r_1r_2}))$ 
%
%
%Pick $n_2=6+14k_0$ and $\tau_1 = \min\{1/2,\sqrt{(8/3)(\mu_0^2r^2d/n_1)\log(8n_2r^2d^{1+\beta})}\}$. 
%By (\ref{tensor-ineq-1}), 
%\bes
%&& \P\{\|(\calQ_{\bT}\calR_{n_2}\calQ_{\bT})\cdots(\calQ_{\bT}\calR_1\calQ_{\bT})\|\ge \tau_1^{n_2}\}
%\\ &\le&\P\{\max_{1\le \ell\le n_2}\|\calQ_{\bT}\calR_{\ell}\calQ_{\bT}\|\ge \tau_1\}\\
%&\le&2n_2r^2 d\exp\Big(-\frac{ \tau_1^2}{8/3}\Big(\frac{n_1}{\mu^2_0 r^2 d}\Big)\Big)\\
%&\le& d^{-\beta}/4. 
%\ees
%Let $r_j=r_j(\bX)$. 
%Since $\|\bW\|_{\rm HS}\le r_1r_2$ by Lemma \ref{lm-alpha}, 
%$P\{$ (\ref{dual-cert-1}) holds $\}\ge 1-d^{-\beta}/4$ when 
%$$
%\sqrt{r_1r_2}\tau_1^{n_2/2} \le \sqrt{r_1r_2}\Big(\frac{8\mu_0^2r^2d\log(C^*d^{4+\beta})}{3n_1}\Big)^{3+7k_0} \le  \frac{1}{4}\sqrt{\frac{n}{d_1d_2d_3}}
%$$
%when $8n_2r^2\le C^*d^3$. 
%For sufficiently large $c_2$, the sample size condition allows $n_1$ satisfying 
%\bes
%\frac{3n_1}{8\log(C^*d^{1+\beta})} \ge \max\Big(2\sqrt{rd_1d_2d_3\log d}, 
%((\mu_0r)\vee \sqrt{r_1r_2})^{1/k_0}\mu_0^2r^2d\Big), 
%\ees
%so that 
%\bes
%\sqrt{r_1r_2}\tau_1^{n_2/2} &\le &  \sqrt{r_1r_2}\Big(\frac{\mu_0^2r^2d}{2\sqrt{rd_1d_2d_3\log d}}\Big)^{3} 
%((\mu_0r)\vee \sqrt{r_1r_2})^{-7}
%\cr &\le &  \frac{d^3/8}{(rd_1d_2d_3\log d)^{3/2}} 
%\cr &\le& \frac{1}{8}\sqrt{\frac{n}{d_1d_2d_3}}. 
%\ees

\medskip
Because
$$
(d_1d_2d_3)^{-1/2}\|\bW\|_{\rm HS}\le \|\bW\|_{\max}\le \|\bW\|\le 1,
$$
%%%% Equation (\ref{dual-cert-1}) follows if
Equation (\ref{dual-cert-1}) holds if $\max_{1\le \ell\le n_2}\|\calQ_{\bT}\calR_{\ell}\calQ_{\bT}\|\le\tau$ and 
\bel{eq:reqn2}
n_2\ge  -{1\over \log \tau}\log\left(\sqrt{32}d_1d_2d_3n^{-1/2}\right).
\eel
Thus, an application of (\ref{tensor-ineq-1}) now gives 
%%%% the probability that (\ref{dual-cert-1}) holds: 
the following bound: 
\bes
\P\Big\{\hbox{ (\ref{dual-cert-1}) holds }\Big\}&\ge & 1-\P\{\|(\calQ_{\bT}\calR_{n_2}\calQ_{\bT})\cdots(\calQ_{\bT}\calR_1\calQ_{\bT})\|\ge \tau^{n_2}\} \\
&\le& 1- \P\{\max_{1\le \ell\le n_2}\|\calQ_{\bT}\calR_{\ell}\calQ_{\bT}\|\ge \tau\}\\
&\le&1- 2n_2r^2 d\exp\Big(-\frac{ \tau^2/2}{1+2\tau /3}\Big(\frac{n_1}{\mu^2_0r^2 d}\Big)\Big).
\ees
%Taking $\tau=1/2$,  then with probability at least
%$$
%1-2n_2\rbar^2(\bT) d\exp\Big(-{2\over 32}\Big(\frac{n_1}{\mu^2(\bT) \rbar^2(\bT)}\Big)\Big),
%$$
%Equation (\ref{dual-cert-1}) holds with the same probability bound as soon as
%$$
%n_2\ge {5\over 2}+{1\over \log 2}\log\left({n^{-1/2}d_1d_2d_3}\right),
%$$
%and in particular, if $n_2=3+\lfloor 5\log d\rfloor$.

Now consider Equation (\ref{dual-cert-2}). 
%%%%
Let $\bW_\ell = \calQ_{\bT}\calR_{\ell}\calQ_{\bT}\bW_{\ell-1}$ for $\ell\ge 1$ with $\bW_0=\bW$. 
Observe that (\ref{dual-cert-2}) does not hold with at most probability 
\bes
&& \P\Big\{\Big\|\sum_{\ell=1}^{n_2} \calR_\ell\bW_{\ell-1}\Big\| \ge 1/4\Big\}\\
&\le& \P\Big\{\big\|\calR_1\bW_{0}\big\| \ge  1/8\Big\} + \P\Big\{\big\|\bW_1\big\|_{\max} \ge  \|\bW\|_{\max}/2 \Big\}
\\ && + \P\Big\{\Big\|\sum_{\ell=2}^{n_2} \calR_\ell\bW_{\ell-1}\Big\| \ge  1/8,\ \big\|\bW_1\big\|_{\max} < \|\bW\|_{\max}/2\Big\} \\
&\le& \P\Big\{\big\|\calR_1\bW_{0}\big\| \ge  1/8\Big\} + \P\Big\{\big\|\bW_1\big\|_{\max} \ge  \|\bW\|_{\max}/2 \Big\}
\\ && + \P\Big\{\big\|\calR_2\bW_{1}\big\| \ge  1/16,\ \big\|\bW_1\big\|_{\max} < \|\bW\|_{\max}/2\Big\} 
\\ && + \P\Big\{\big\|\bW_2\big\|_{\max} \ge  \|\bW\|_{\max}/4,\ \big\|\bW_1\big\|_{\max} < \|\bW\|_{\max}/2 \Big\}
\\ && + \P\Big\{\Big\|\sum_{\ell=3}^{n_2} \calR_\ell\bW_{\ell-1}\Big\| \ge  {1\over 16},\ 
\big\|\bW_2\big\|_{\max} < \|\bW\|_{\max}/4\Big\} \\
&\le &\sum_{\ell=1}^{n_2-1}
\P\Big\{\big\|\calQ_{\bT}\calR_{\ell}\calQ_{\bT}\bW_{\ell-1}\big\|_{\max} \ge  \|\bW\|_{\max}/2^{\ell},\ 
\big\|\bW_{\ell-1}\big\|_{\max} \le \|\bW\|_{\max}/2^{\ell-1}\Big\} \\
&& + \sum_{\ell=1}^{n_2}
\P\Big\{\big\|\calR_\ell\bW_{\ell-1}\big\| \ge  2^{-2-\ell},\ \big\|\bW_{\ell-1}\big\|_{\max} \le \|\bW\|_{\max}/2^{\ell-1}\Big\}. 
\ees
Since $\{\calR_\ell,\bW_\ell\}$ are i.i.d., (\ref{tensor-ineq-2}) with 
$\bX= \bW_{\ell-1}/\|\bW_{\ell-1}\|_{\max}$ implies 
\bes
&& \P\Big\{\hbox{ (\ref{dual-cert-2}) holds }\Big\} \\
%&& \P\Big\{\Big\|\sum_{\ell=1}^{n_2} \calR_\ell\bW_{\ell-1}\Big\| > 1/4\Big\}\\
&\ge& 1-n_2\max_{\substack{\bX: \bX=\calQ_{\bT}\bX\\ \|\bX\|_{\max}\le 1}}
\left(\P\Big\{\Big\|\calQ_{\bT}\calR_1\calQ_{\bT}\bX\Big\|_{\max} > {1\over 2}\Big\}
+\P\Big\{\Big\|\calR_1\bX\Big\| > {1\over 8\|\bW\|_{\max}}\Big\}
\right)\\
&\ge& 1-2n_2d_1d_2d_3\exp\Big(\frac{-(3/32)n_1}{\mu_0^2 r^2 d}\Big)
- n_2\max_{\substack{\bX: \bX=\calQ_{\bT}\bX\\ \|\bX\|_{\max}\le \|\bW\|_{\max}}}
\P\Big\{\Big\|\calR_1\bX\Big\| > {1\over 8}\Big\}. 
\ees
The last term on the right hand side can be bounded using the following result.

\vskip 20pt
\begin{lemma}\label{lm-tensor-sparse-ineq}
Assume that $\alpha(\bT)\le \alpha_0$, $\rbar(\bT)=r$ and 
$q^*_1=  \big(\beta + \log d\big)^2\alpha_0^2r\log d$.  
There exists a numerical constant $c_1>0$ such that for 
any constants $\beta>0$ and $1/(\log d)\le \delta_1 <1$, 
\bel{eq:asydim}
n_1 \ge c_1\left[q^*_1d^{1+\delta_1}
+ \sqrt{q^*_1(1+\beta)\delta_1^{-1}d_1d_2d_3}\right]
\eel
implies 
\bel{eq:spabd}
\max_{\substack{\bX: \bX=\calQ_{\bT}\bX\\ \|\bX\|_{\max}\le \|\bW\|_{\max}}}
\P\Big\{\Big\|\calR_1\bX\Big\| \ge {1\over 8}\Big\}\le d^{-\beta-1},
\eel
where $\bW$ is in the range of $\calQ_{\bT}^0$ such that $\|\bW\|=1$ and $\langle \bT,\bW\rangle=\|\bT\|_\ast$. 
\end{lemma}

\subsection{Proof of Theorem \ref{th-1}} 
Since (\ref{dual-cert}) is a consequence of (\ref{dual-cert-1}) and (\ref{dual-cert-2}), 
it follows from Lemmas \ref{prop-dual-cert}, \ref{lm:op-norm}, \ref{lm-tensor-ineq} and \ref{lm-tensor-sparse-ineq} 
that for $\tau\in (0,1/2]$ and $n\ge n_1n_2$ satisfying conditions (\ref{eq:reqn2}) and (\ref{eq:asydim}), 
%the conditions on $n_1$ in Lemma \ref{lm-tensor-sparse-ineq} 
\bes
\P\Big\{\hbT\neq \bT\Big\}
&\le& 2r^2 d\,\exp\Big(-\frac{3}{32}\Big(\frac{n}{\mu^2_0 r^2 d}\Big)\Big) 
+ 2n_2r^2 d\exp\Big(-\frac{ \tau^2/2}{1+2\tau /3}\Big(\frac{n_1}{\mu^2_0r^2 d}\Big)\Big)
\cr && + 2n_2d_1d_2d_3\exp\Big(\frac{-(3/32)n_1}{\mu_0^2 r^2 d}\Big)
+ n_2 d^{-\beta-1}. 
\ees
We now prove Theorem \ref{th-1} by setting $\tau = d^{-\delta_2/2}/2$, so that condition 
(\ref{eq:reqn2}) can be written as $n_2\ge c_2/\delta_2$. 
Assume without loss of generality $n_2\le d/2$ because large $c_0$ forces large $d$. 
For sufficiently large $c_2'$, 
the right-hand side of the above inequality is no greater than $d^{-\beta}$ when 
\bes
n_1 \ge c_2'(1+\beta)(\log d)\mu_0^2r^2 d/(4\tau^2) = c_2' q^*_2 d^{1+\delta_2}
\ees
holds as well as (\ref{eq:asydim}). Thus, (\ref{case-1}) implies (\ref{eq:rec}) for sufficiently large $c_0$. 
$\hfill\square$

\section{Concentration Inequalities for Low Rank Tensors}
\label{sec:low}

We now prove Lemmas \ref{lm:op-norm} and \ref{lm-tensor-ineq}, both involving tensors of low rank. We note that Lemma \ref{lm:op-norm} concerns the concentration inequality for the sum of a sequence of dependent tensors whereas in Lemma \ref{lm-tensor-ineq}, we are interested in a sequence of iid tensors.

\subsection{Proof of Lemma \ref{lm:op-norm}}
We first consider Lemma \ref{lm:op-norm}. Let $(a_k,b_k,c_k)$ be sequentially uniformly sampled from $\Omega^*$ without replacement, $S_k = \{(a_j,b_j,c_j): j\le k\}$, and $m_k=d_1d_2d_3-k$. Given $S_k$, the conditional expectation of $\calP_{(a_{k+1},b_{k+1},c_{k+1})}$ is 
$$
\E\Big[\calP_{(a_{k+1},b_{k+1},c_{k+1})}\Big|S_k\Big]= \frac{\calP_{S_k^c}}{m_k}. 
$$
For $k=1,\ldots,n$, define martingale differences 
$$
\calD_{k} = d_1d_2d_3(m_n/m_k)\calQ_{\bT}\Big(\calP_{(a_{k},b_{k},c_{k})} 
- \calP_{S_{k-1}^c}/m_{k-1}\Big)\calQ_{\bT}. 
$$
Because $\calP_{S_0^c}=\calI$ and $S_n=\Omega$, we have 
\bes
\calQ_{\bT}\calP_{\Omega}\calQ_{\bT}/m_n
&=& \frac{\calD_n}{d_1d_2d_3m_n} 
+ \calQ_{\bT}(\calP_{S_{n-1}^c}/m_{n-1})\calQ_{\bT}/m_n
+\calQ_{\bT}\calP_{S_{n-1}}\calQ_{\bT}/m_n
\cr &=& \frac{\calD_n}{d_1d_2d_3m_n} + \calQ_{\bT}(1/m_n-1/m_{n-1})
+\calQ_{\bT}\calP_{S_{n-1}}\calQ_{\bT}/m_{n-1}
\cr &=& \sum_{k=1}^n\frac{\calD_k}{d_1d_2d_3m_n} +\calQ_{\bT}(1/m_n-1/m_0). 
\ees
Since $1/m_n-1/m_0 = n/(d_1d_2d_3m_n)$, it follows that 
$$
\calQ_{\bT}(d_1d_2d_3/n)\calP_{\Omega}\calQ_{\bT} - \calQ_{\bT}= \frac{1}{n}\sum_{k=1}^{n} \calD_k. 
$$
Now an application of the matrix martingale Bernstein inequality (see, e.g., Tropp, 2011) gives 
\bes
\P\Big\{\frac{1}{n}\Big\|\sum_{k=1}^n\calD_k\Big\|> \tau \Big\}\le 2\,\rank(\calQ_{\bT})\exp\Big(\frac{ - n^2\tau^2/2}{\sigma^2+ n\tau M/3}\Big), 
\ees
where $M$ is a constant upper bound of $\|\calD_k\|$ and $\sigma^2$ is a constant upper bound of
$$\big\|\sum_{k=1}^n \E\big[ \calD_k\calD_k|S_{k-1}\big]\big\|.$$ 
Note that $D_k$ are random self-adjoint operators. 

Recall that $\calQ_{\bT}$ can be decomposed as a sum of orthogonal projections 
\bes
\calQ_{\bT} &=& (\calQ_{\bT}^0+\calQ_{\bT}^1)+\calQ_{\bT}^2+\calQ_{\bT}^3
\cr &=& \bI\otimes \bP_{\bT}^2 \otimes \bP_{\bT}^3 
+\bP_{\bT}^1\otimes \bP_{\bT^\perp}^2\otimes \bP_{\bT}^3
+ \bP_{\bT}^1\otimes \bP_{\bT}^2\otimes \bP_{\bT^\perp}^3. 
\ees
The rank of $\calQ_{\bT}$, or equivalently the dimension of its range, is given by 
\bes
d_1r_2r_3+(d_2-r_2)r_1r_3+(d_3-r_3)r_1r_2 \le \rbar^2 d.
\ees
Hereafter, we shall write $r_j$ for $r_j(\bT)$, $\mu$ for $\mu(\bT)$, and $\rbar$ for $\rbar(\bT)$ for brevity when no confusion occurs. Since $\E\big[\calD_k\big|S_{k-1}\big]=0$, the total variation is bounded by 
\bes
&& \max_{\|\calQ_{\bT}\bX\|_{\rm HS}=1}\sum_{k=1}^n 
\E\Big[\Big\langle \calD_k\bX,\calD_k\bX\Big\rangle\Big|S_{k-1}\Big]
\cr &\le & \max_{\|\calQ_{\bT}\bX\|_{\rm HS}=1}\sum_{k=1}^n
\Big(d_1d_2d_3(m_n/m_k)\Big)^2
\E\Big[\Big\langle \big(\calQ_{\bT}\calP_{(a_{k},b_{k},c_{k})}\calQ_{\bT}\big)^2\bX,\bX\Big\rangle\Big|S_{k-1}\Big]
\cr &\le& \sum_{k=1}^n
\Big(d_1d_2d_3(m_n/m_k)\Big)^2 m^{-1}_{k-1}
\max_{\|\calQ_{\bT}\bX\|_{\rm HS}=1}\sum_{a,b,c}
\Big\langle \big(\calQ_{\bT}\calP_{(a,b,c)}\calQ_{\bT}\big)^2\bX,\bX\Big\rangle. 
\ees
Since $m_n\le m_k$ and $\sum_{k=1}^n (m_n/m_k)/m_{k-1}=n/d_1d_2d_3$, 
\bes
\max_{\|\calQ_{\bT}\bX\|_{\rm HS}=1}\sum_{k=1}^n 
\E\Big[\Big\langle \calD_k\bX,\calD_k\bX\Big\rangle\Big|S_{k-1}\Big]
\le nd_1d_2d_3\max_{a,b,c}\big\|\calQ_{\bT}\calP_{(a,b,c)}\calQ_{\bT}\big\|. 
\ees
It then follows that 
\bes
\max_{a,b,c}\big\|\calQ_{\bT}\calP_{(a,b,c)}\calQ_{\bT}\big\|
&=& \max_{\|\calQ_{\bT}\bX\|_{\rm HS}=1}\Big\langle \calQ_{\bT}\calP_{(a,b,c)}\calQ_{\bT}\bX,\calQ_{\bT}\bX\Big\rangle 
\cr &=& \max_{\|\calQ_{\bT}\bX\|_{\rm HS}=1}\Big\langle \calQ_{\bT}\bfe_a\otimes \bfe_b\otimes \bfe_c ,\calQ_{\bT}\bX\Big\rangle^2
\cr &\le & \frac{\mu^2 \rbar^2 d}{d_1d_2d_3}. 
\ees
Consequently, we may take $\sigma^2 = n\mu^2_0 \rbar^2 d$. Similarly, 
$$
M\le \max_k d_1d_2d_3(m_n/m_k)2\max_{a,b,c}\big\|\calQ_{\bT}\calP_{(a,b,c)}\calQ_{\bT}\big\|
\le 2\mu^2 \rbar^2 d. 
$$
Inserting the expression and bounds for $\rank(\calQ_{\bT})$, $\sigma^2$ and $M$ into the 
Bernstein inequality, we find 
$$
\P\Big\{\frac{1}{n}\Big\|\sum_{k=1}^n\calD_k\Big\| > \tau \Big\}
\le 2(\rbar^2 d)\exp\Big(\frac{ - \tau^2/2}{1+2\tau /3}\Big(\frac{n}{\mu^2 \rbar^2 d}\Big)\Big),
$$
which completes the proof because $\mu(\bT)\le \mu_0$ and $\rbar(\bT)=r$.$\hfill\square$

\subsection{Proof of Lemma \ref{lm-tensor-ineq}.}
In proving Lemma \ref{lm-tensor-ineq}, we consider first (\ref{tensor-ineq-2}). Let $\bX$ be a tensor with $\|\bX\|_{\max}\le 1$. Similar to before, write
$$
\calD_i=d_1d_2d_3\calQ_{\bT}\calP_{(a_i,b_i,c_i)}-\calQ_{\bT}
$$
for $i=1,\ldots,n_1$. Again, we shall also write $\mu$ for $\mu(\bT)$, and $\rbar$ for $\rbar(\bT)$ for brevity. Observe that for each point $(a,b,c)\in[d_1]\times[d_2]\times[d_3]$, 
\bes
&& {1\over d_1d_2d_3}\big|\langle \bfe_a\otimes \bfe_b\otimes \bfe_c ,\calD_i \bX \rangle\big|
\cr &=&\Big| \langle \calQ_{\bT}(\bfe_a\otimes \bfe_b\otimes \bfe_c) ,\calQ_{\bT}(\bfe_{a_k}\otimes\bfe_{b_k}\otimes\bfe_{c_k})\rangle X(a_k,b_k,c_k) 
- \langle \calQ_{\bT}(\bfe_a\otimes \bfe_b\otimes \bfe_c) ,\calQ_{\bT}\bX\rangle/(d_1d_2d_3)\Big| 
\cr &\le& 2\max_{a,b,c}\|\calQ_{\bT}(\bfe_a\otimes \bfe_b\otimes \bfe_c) \|_{\rm HS}^2\|\bX\|_{\max}
\cr &\le& 2 \mu^2 \rbar^2 d/(d_1d_2d_3). 
\ees
Since the variance of a variable is no greater than the second moment,  
\bes
&& \E\left({1\over d_1d_2d_3}\big|\langle \bfe_a\otimes \bfe_b\otimes \bfe_c ,\calQ_{\bT}\calD_i \bX \rangle\big|\right)^2
\cr &\le& \E \big| \langle \calQ_{\bT}(\bfe_a\otimes \bfe_b\otimes \bfe_c) ,\calQ_{\bT}(\bfe_{a'}\otimes \bfe_{b'}\otimes \bfe_{c'})\rangle X(a_k,b_k,c_k)\big|^2
\cr &\le& {1\over d_1d_2d_3}\sum_{a',b',c'}\big| \langle \calQ_{\bT}\bfe_a\otimes \bfe_b\otimes \bfe_c ,\calQ_{\bT}(\bfe_{a'}\otimes \bfe_{b'}\otimes \bfe_{c'})\rangle\big|^2
\cr & = & {1\over d_1d_2d_3}\|\calQ_{\bT}(\bfe_a\otimes \bfe_b\otimes \bfe_c) \|_{\rm HS}^2
\cr &\le& \mu^2 \rbar^2 d/(d_1d_2d_3)^2. 
\ees
Since $\langle \bfe_a\otimes \bfe_b\otimes \bfe_c ,\calQ_{\bT}\calD_i \bX \rangle$ are iid random variables, the Bernstein inequality yields
\bes
\P\Big\{\Big|\frac{1}{n_1}\sum_{i=1}^{n_1} \langle \bfe_a\otimes \bfe_b\otimes \bfe_c ,\calQ_{\bT}\calD_i \bX \rangle\Big| > \tau\Big\}
\le 2\exp\Big( - \frac{(n_1\tau)^2/2}{n_1\mu^2 \rbar^2 d+n_1\tau 2\mu^2 \rbar^2 d/3}\Big). 
\ees
This yields (\ref{tensor-ineq-2}) by the union bound. 

The proof of (\ref{tensor-ineq-1}) is similar, but the matrix Bernstein inequality is used. We equip $\R^{d_1\times d_2\times d_3}$ with the Hilbert-Schmidt norm so that 
it can be viewed as the Euclidean space. As linear maps in this Euclidean space, the operators $D_i$ are just random matrices. Since the projection 
$\calP_{(a,b,c)}: \bX \to \langle \bfe_a\otimes \bfe_b\otimes \bfe_c , \bX\rangle \bfe_a\otimes \bfe_b\otimes \bfe_c $ is of rank 1, 
\bes
\|\calQ_{\bT}\calP_{(a,b,c)}\calQ_{\bT}\| = \|\calQ_{\bT}(\bfe_a\otimes \bfe_b\otimes \bfe_c) \|_{\rm HS}^2\le \mu^2 \rbar^2 d/(d_1d_2d_3). 
\ees
It follows that $\|\calD_i\| \le 2\mu^2 \rbar^2 d$. Moreover, $\calD_i$ is a self-adjoint operator and its covariance operator is bounded by 
\bes
\max_{\|\bX\|_{\rm HS}=1}\E\big\|\calD_i\bX\|_{\rm HS}^2
&\le& (d_1d_2d_3)^2\max_{\|\bX\|_{\rm HS}=1} 
\E\Big\langle \big(\calQ_{\bT}\calP_{(a_{k},b_{k},c_{k})}\calQ_{\bT}\big)^2\bX,\bX\Big\rangle
\cr &\le& d_1d_2d_3\sum_{a,b,c}\|\calQ_{\bT}(\bfe_a\otimes \bfe_b\otimes \bfe_c) \|_{\rm HS}^2
\big\langle \bfe_a\otimes \bfe_b\otimes \bfe_c ,\bX\big\rangle^2 
\cr & \le & d_1d_2d_3\max_{a,b,c}\|\calQ_{\bT}(\bfe_a\otimes \bfe_b\otimes \bfe_c) \|_{\rm HS}^2
\cr & = & \mu^2 \rbar^2 d
\ees
Consequently, by the matrix Bernstein inequality (Tropp, 2011),
\bes
\P\Big\{\Big\|\frac{1}{n_1}\sum_{i=1}^{n_1} \calD_i\Big\| > \tau \Big\}
\le 2\,\rank(\calQ_{\bT}) \exp\Big(\frac{\tau^2/2}{1 + 2\tau/3}\Big(\frac{n_1}{\mu^2 \rbar^2 d}\Big)\Big). 
\ees
This completes the proof due to the fact that $\rank(\calQ_{\bT})\le\rbar^2 d$. $\hfill\square$ 

\section{Concentration Inequalities for Sparse Tensors}
\label{sec:sparse}

%%%%
We now derive probabilistic bounds for $\|\calR_\ell\bX\|$ %in Lemma \ref{lm-tensor-sparse-ineq} 
when $(a_i,b_i,c_i)$s are iid vectors uniformly sampled from $[d_1]\times[d_2]\times[d_3]$ 
and $\bX = \calQ_{\bT}\bX$ with small $\|\bX\|_{\max}$. 

\subsection{Symmetrization}
We are interested in bounding
$$
\max_{\bX\in\scrU(\eta)}\P\left\{\left\|{d_1d_2d_3\over n}\sum_{i=1}^n \calP_{(a_i,b_i,c_i)}\bX-\bX\right\|\ge t\right\}, 
$$
%%%%
e.g. with $(n,\eta,t)$ replaced by $(n_1,(\alpha_0\sqrt{r})\wedge\sqrt{d_1d_2d_3},1/8)$ in the proof of 
Lemma \ref{lm-tensor-sparse-ineq}, where 
\bes
\scrU(\eta) = \{\bX: \calQ_{\bT}\bX=\bX, \|\bX\|_{\max}\le \eta/\sqrt{d_1d_2d_3}\}. 
\ees
Our first step is symmetrization. 

\begin{lemma}\label{lm-sym} 
Let $\epsilon_i$s be a Rademacher sequence, that is a sequence of i.i.d. $\eps_i$ with 
$\P\{\epsilon_i=1\}=\P\{\epsilon_i=-1\}=1/2$. Then
\bes
&&\max_{\bX\in\scrU(\eta)}\P\left\{\left\|{d_1d_2d_3\over n}\sum_{i=1}^n \calP_{(a_i,b_i,c_i)}\bX-\bX\right\|\ge t\right\}\\
&\le& 4\max_{\bX\in\scrU(\eta)}\P\left\{\left\|{d_1d_2d_3\over n}\sum_{i=1}^n \epsilon_i\calP_{(a_i,b_i,c_i)}\bX\right\|\ge t/2\right\}+4\exp\left(-{nt^2/2\over \eta^2+2\eta t\sqrt{d_1d_2d_3}/3}\right).
\ees

\end{lemma}

\vskip 20pt
\noindent {\sc Proof of Lemma \ref{lm-sym}.} 
The standard symmetrization argument gives 
\bes
&&\max_{\bX\in\scrU(\eta)}\P\left\{\left\|{d_1d_2d_3\over n}\sum_{i=1}^n \calP_{(a_i,b_i,c_i)}\bX-\bX\right\|\ge t\right\}\\
&\le& 4\max_{\bX\in\scrU(\eta)}\P\left\{\left\|{d_1d_2d_3\over n}\sum_{i=1}^n \epsilon_i\calP_{(a_i,b_i,c_i)}\bX\right\|\ge t/2\right\}\\
&&\hskip 10pt +\,2\max_{\bX\in\scrU(\eta)}
\max_{\|\bu\|=\|\bv\|=\|\bw\|=1}\P\Big\{
\Big\langle \bu\otimes\bv\otimes\bw, {d_1d_2d_3\over n}\sum_{i=1}^n \calP_{(a_i,b_i,c_i)}\bX-\bX\Big\rangle > t\Big\}.  
\ees
It remains to bound the second quantity on the right-hand side. To this end, denote by
$$\xi_i=\big\langle \bu\otimes\bv\otimes\bw, d_1d_2d_3\calP_{(a_i,b_i,c_i)}\bX-\bX\big\rangle.$$
For $\|\bu\|=\|\bv\|=\|\bw\|=1$ and $\bX\in\scrU(\eta)$, $\xi_i$ are iid variables with $\E\xi_i=0$, 
$|\xi_i|\le 2d_1d_2d_3\|\bX\|_{\max} \le 2\eta\sqrt{d_1d_2d_3}$ and 
$\E\xi_i^2\le (d_1d_2d_3)\|\bX\|_{\max}^2 \le \eta^2$. 
Thus, the statement follows from the Bernstein inequality.$\hfill\square$ 

%$$\calD_i=(d_1d_2d_3)\calP_{(a_i,b_i,c_i)}-\calI.$$
%Then for $\|\bu\|=\|\bv\|=\|\bw\|=1$ and $\bX\in\scrU(\eta)$, $\big\langle \bu\otimes\bv\otimes\bw,\calD_i\bX\big\rangle$ are iid variables with
%\bes
%\E \big\langle \bu\otimes\bv\otimes\bw,\calD_i\bX\big\rangle &=& 0;
%\cr \big|\big\langle \bu\otimes\bv\otimes\bw,\calD_i\bX\big\rangle\big|
%&\le&  2d_1d_2d_3\|\bX\|_{\max} \le 2\eta\sqrt{d_1d_2d_3};
%\cr \E \big\langle \bu\otimes\bv\otimes\bw,\calD_i\bX\big\rangle ^2
%&\le & \E \big\langle \bu\otimes\bv\otimes\bw,(d_1d_2d_3)\calP_{(a_i,b_i,c_i)}\bX\big\rangle^2
%\cr &\le& (d_1d_2d_3)\|\bX\|_{\max}^2 \le \eta^2. 
%\ees
%Thus, by Bernstein's inequality,
%\bes
%\P\Big\{
%\Big\langle \bu\otimes\bv\otimes\bw, {d_1d_2d_3\over n}\sum_{i=1}^n \calP_{(a_i,b_i,c_i)}\bX-\bX\Big\rangle > t\Big\}\le 2\exp\left(-{nt^2/2\over \eta^2+2\eta t\sqrt{d_1d_2d_3}/3}\right),
%\ees
%and the statement then follows.$\hfill\square$ 
%\vskip 20pt

In the light of Lemma \ref{lm-sym}, it suffices to consider bounding
$$
\max_{\bX\in\scrU(\eta)}\P\left\{\left\|{d_1d_2d_3\over n}\sum_{i=1}^n \epsilon_i\calP_{(a_i,b_i,c_i)}\bX\right\|\ge t/2\right\}.
$$
To this end, we use a thinning method to control the spectral norm of tensors.

%%%%%%%%%%%%%%%%%%%%%%%%%%%%%
\subsection{Thinning of the spectral norm of tensors}
Recall that the spectral norm of a tensor $\bZ\in \R^{d_1\times d_2\times d_3}$ is defined as 
$$
\|\bZ\| = \max_{\substack{\bu\in \R^{d_1}, \bv\in \R^{d_2}, \bw\in \R^{d_3}\\\|\bu\|\vee \|\bv\|\vee \|\bw\|\le 1}} \langle \bu\otimes \bv\otimes \bw, \bZ\rangle. 
$$
We first use a thinning method to discretize maximization in the unit ball in $\R^{d_j}$ to the problem involving only vectors taking values $0$ or $\pm 2^{-\ell/2}, \ell\le m_j:=\lceil \log_2 d_j\rceil$, that is, binary ``digitalized'' vectors that belong to 
\bel{class-B}
\scrB_{m_j,d_j} &=& \{0,\pm1,\pm 2^{-1/2},\ldots,\pm 2^{-m_j/2}\}^d\cap\{\bu\in \R^{d_j}: \|\bu\|\le 1\}. 
\eel

\begin{lemma}\label{lm-thinning} 
For any tensor $\bZ\in \R^{d_1\times d_2\times d_3}$,
$$
\|\bZ\| \le 8\max\Big\{\langle \bu\otimes \bv\otimes \bw, \bZ\rangle: \bu\in \scrB_{m_1,d_1},\bv\in \scrB_{m_2,d_2},\bw\in \scrB_{m_3,d_3}\Big\},
$$
where $m_j:=\lceil \log_2 d_j\rceil$, $j=1,2,3$.
\end{lemma}

\vskip 20pt
\noindent {\sc Proof of Lemma \ref{lm-thinning}.}
Denote by  
\bes
C_{m,d} = \min_{\|\ba\|=1}\max_{u\in \scrB_{m,d}}\bu^\top\ba,
\ees
which bounds the effect of discretization. Let $\bX$ be a linear mapping from $\R^d$ to a linear space equipped with a seminorm $\|\cdot\|$. Then, $\|\bX \bu\|$ can be written as the maximum of $\phi(\bX\bu)$ over linear functionals $\phi(\cdot)$ of unit dual norm. Since $\max_{\|\bu\|\le 1}\bu^\top\ba=1$ for $\|{\ba}\|=1$, it follows from the definition of $C_{m,d}$ that 
$$
\max_{\|\bu\|\le 1}\bu^\top\ba \le \|\ba\|C_{m,d}^{-1} \max_{\bu\in \scrB_{m,d}}\bu^\top(\ba/\|\ba\|)=C_{m,d}^{-1} \max_{\bu\in \scrB_{m,d}}\bu^\top\ba
$$ 
for every ${\ba}\in \R^d$ with $\|\ba\|>0$. Consequently, for any positive integer $m$, 
$$
\max_{\|\bu\|\le 1} \|\bX\bu\| = \max_{\ba:\ba^\top\bv = \phi(\bX\bv)\forall\bv}\max_{\|\bu\|\le 1}\ba^\top\bu 
\le C_{m,d}^{-1}\max_{\bu\in \scrB_{m,d}} \|\bX\bu\|. 
$$
An application of the above inequality to each coordinate yields
$$
\|\bZ\| \le C_{m_1,d_1}^{-1}C_{m_2,d_2}^{-1}C_{m_3,d_3}^{-1}
\max_{\bu\in \scrB_{m_1,d_1},\bv\in \scrB_{m_2,d_2},\bw\in \scrB_{m_3,d_3}}
\langle  \bu\otimes \bv\otimes \bw,\bZ\rangle. 
$$

It remains to show that $C_{m_j,d_j}\ge 1/2$. To this end, we prove a stronger result that for any $m$ and $d$,
$$
C_{m,d}^{-1}\le \sqrt{2+2(d-1)/(2^m-1)}.
$$
Consider first a continuous version of $C_{m,d}$:
\bes
C_{m,d}' = \min_{\|{\ba}\|=1}\max_{\bu\in \scrB_{m,d}'}\ba^\top \bu. 
\ees
where $\scrB_{m,d}'=\{t: t^2\in [0,1]\setminus (0,2^{-m})\}^d\cap\{\bu: \|\bu\|\le 1\}$. 
Without loss of generality, we confine the calculation to nonnegative ordered ${\ba}=({a}_1,\ldots,{a}_d)^\top$ satisfying $0\le {a}_1\le \ldots \le {a}_d$ and $\|{\ba}\|=1$. Let 
$$
k = \max\Big\{j: 2^m {a}_j^2 + \sum_{i=1}^{j-1} {a}_i^2 \le 1\Big\}\quad {\rm and}\quad  \bv = \frac{({a}_iI\{i>k\})_{d\times 1}}{\{1-\sum_{i=1}^{k} {a}_i^2\}^{1/2}}. 
$$
Because $2^m v_{k+1}^2 = 2^m{a}_{k+1}^2/(1-\sum_{i=1}^{k} {a}_i^2)\ge 1$, 
we have $\bv\in \scrB_{m,d}'$. By the definition of $k$, there exists $x^2\ge {a}_k^2$ satisfying 
$$(2^m-1)x^2 + \sum_{i=1}^{k} {a}_i^2 = 1.$$
It follows that 
$$
\sum_{i=1}^k {a}_i^2 = \frac{\sum_{i=1}^k {a}_i^2}{(2^m-1)x^2+\sum_{i=1}^k {a}_i^2} \le \frac{kx^2}{(2^m-1)x^2+kx^2} \le \frac{d-1}{2^m+d-2}. 
$$
Because $\ba^\top \bv = (1-\sum_{i=1}^{k} {a}_i^2)^{1/2}$ for this specific $\bv\in \scrB_{m,d}'$, we get
$$
C_{m,d}' \ge \min_{\|{a}\|_2=1}\Big(1-\sum_{i=1}^{k} {a}_i^2\Big)^{1/2}
\ge \Big(1-\frac{d-1}{2^m+d-2}\Big)^{1/2}=\Big(\frac{2^m-1}{2^m+d-2}\Big)^{1/2}.
$$ 
Now because every $\bv\in \scrB_{m,d}'$ with nonnegative components matches a $\bu\in \scrB_{m,d}$ with 
$$\sign(v_i)\sqrt{2}u_i\ge |v_i|\ge \sign(v_i)u_i,$$
we find $C_{m,d}\ge C'_{m,d}/\sqrt{2}$. Consequently, 
$$1/C_{m,d}\le \sqrt{2}/C'_{m,d}\le \sqrt{2}\{1+(d-1)/(2^m-1)\}^{1/2}.\hskip 50pt\square$$
\vskip 20pt

It follows from Lemma \ref{lm-thinning} that the spectrum norm $\|\bZ\|$ is of the same order as the maximum of 
$\langle \bu\otimes \bv\otimes \bw, \bZ\rangle$ over $\bu\in \scrB_{m_1,d_1}$, 
$\bv\in\scrB_{m_2,d_2}$ and $\bw\in\scrB_{m_3,d_3}$. 
We will further decompose such tensors $\bu\otimes \bv\otimes \bw$ according to the absolute value 
of their entries and bound the entropy of the components in this decomposition. 

\subsection{Spectral norm of tensors with sparse support}
Denote by $D_j$ a ``digitalization'' operator such that $D_j(\bX)$ will zero out all entries of $\bX$ 
whose absolute value is not $2^{-j/2}$, that is
\bel{D_j}
D_j(\bX) = \sum_{a,b,c} 
{\mathbb I}\big\{|\langle \bfe_a\otimes \bfe_b\otimes \bfe_c ,\bX\rangle| = 2^{-j/2}\big\}\langle \bfe_a\otimes \bfe_b\otimes \bfe_c ,\bX\rangle \bfe_a\otimes \bfe_b\otimes \bfe_c . 
\eel
With this notation, it is clear that for $\bu\in \scrB_{m_1,d_1}$, $\bv\in \scrB_{m_2,d_2}$ and $\bw\in \scrB_{m_3,d_3}$, 
$$
\langle \bu\otimes \bv\otimes \bw, \bX\rangle 
=\sum_{j=0}^{m_1+m_2+m_3}\langle D_j(\bu\otimes \bv\otimes \bw), \bX\rangle. 
$$

The possible choice of $D_j(\bu\otimes \bv\otimes \bw)$ in the above expression may be further reduced if $\bX$ is sparse. More specifically, denote by
$$\supp(\bX)=\{\omega\in [d_1]\times[d_2]\times[d_3]: X(\omega)\neq 0\}.$$
Define the maximum aspect ratio of $\supp(\bX)$ as 
\bel{column-row-bd}
\nu_{\supp(\bX)} =\max_{\ell=1,2,3}\max_{i_k: k\neq \ell} \left|\{i_\ell: (i_1,i_2,i_3)\in \supp(\bX)\}\right|.
\eel
In other words, the quantity $\nu_{\supp(\bX)}$ is the maximum $\ell_0$ norm of the fibers of the third-order tensor. We observe first that, if $\supp(\bX)$ is a uniformly sampled subset of $[d_1]\times [d_2]\times [d_3]$, then it necessarily has a small aspect ratio.

\begin{lemma}
\label{le:aspbd}
Let $\Omega$ be a uniformly sampled subset of $[d_1]\times [d_2]\times [d_3]$ without replacement. 
Let $d=d_1+d_2+d_3$, $p^* = \max(d_1,d_2,d_3)/(d_1d_2d_3)$, 
and $\nu_1 = (d^{\delta_1} enp^*)\vee \{(3+\beta)/\delta_1\}$ with a certain $\delta_1\in [1/\log d,1]$. Then, 
$$
\P\Big\{\nu_\Omega \ge \nu_1\Big\}\le d^{-\beta-1}/3. 
$$
\end{lemma}

\vskip 20pt
\noindent {\sc Proof of Lemma \ref{le:aspbd}.} 
Let $p_1=d_1/(d_1d_2d_3)$, $t = \log(\nu_1/(np^*))\ge 1$, and 
$$N_{i_2i_3}=\left|\{i_\ell: (i_1,i_2,i_3)\in \Omega\}\right|.$$
Because $N_{i_2i_3}$ follows the Hypergeometric$(d_1d_2d_3,d_1,n)$ distribution, its moment generating function is no greater than that of Binomial$(n,p_1)$. Due to $p_1\le p^*$, 
$$
\P\{N_{i_2i_3}\ge \nu_1\} \le \exp\left(-t\nu_1 + np^*(e^t-1)\right)
\le \exp\left(- \nu_1 \log(\nu_1/(enp^*))\right). 
$$
The condition on $\nu_1$ implies $\nu_1 \log(\nu_1/(enp^*)) \ge (3+\beta)\log d$. 
By the union bound, 
$$
\P\{\max_{i_2,i_3}N_{i_2i_3}\ge \nu_1\}\le d_2d_3 d^{-3-\beta}. 
$$
By symmetry, the same tail probability bound also holds for 
$\max_{i_1,i_3}\left|\{i_2: (i_1,i_2,i_3)\in \Omega\}\right|$ 
and $\max_{i_1,i_2}\left|\{i_3: (i_1,i_2,i_3)\in \Omega\}\right|$, so that 
$\P\{\nu_\Omega\ge \nu_1\}\le (d_1d_2+d_1d_3+d_2d_3)d^{-3-\beta}$. 
The conclusion follows from $d_1d_2+d_1d_3+d_2d_3\le d^2/3$. $\hfill\square$
\vskip 20pt

We are now in position to further reduce the set of maximization in 
%%%
defining the spectrum norm of sparse tensors. 
To this end, denote for a block $A\times B\times C\subseteq [d_1]\times [d_2]\times [d_3]$, 
\bel{aspect-ratio}
h(A\times B\times C) = \min\Big\{\nu: |A| \le \nu|B||C|,\ |B|\le \nu|A||C|,\ |C|\le \nu|A||B|\Big\}. 
\eel
It is clear that for any block $A\times B\times C$, there exists $\Atil\subseteq A$, $\Btil\subseteq B$ and $\Ctil\subseteq C$ such that $h(\Atil\times\Btil\times\Ctil)\le \nu_\Omega$ and
$$
(A\times B\times C)\cap\Omega = (\Atil\times \Btil \times \Ctil)\cap \Omega.
$$
For $\bu\in \scrB_{m_1,d_1}$, $\bv\in \scrB_{m_2,d_2}$ and $\bw\in \scrB_{m_3,d_3}$, 
let $A_{i_1} = \{a: u_a^2=2^{-i_1}\}$, $B_{i_2}= \{b: v_b^2=2^{-i_2}\}$ and $C_{i_3}= \{c: w_c^2=2^{-i_3}\}$, 
and define 
\bel{Dtil}
\Dtil_j(\bu\otimes \bv\otimes \bw) = \sum_{(i_1,i_2,i_3):i_1+i_2+i_3=j}
\calP_{\Atil_{j,i_1}\times \Btil_{j,i_2}\times \Ctil_{j,i_3}}D_j(\bu\otimes \bv\otimes \bw)
\eel
where $\Atil_{j,i_1}\subseteq A_{j,i_1}$, $\Btil_{j,i_2}\subseteq B_{j,i_2}$ and $\Ctil_{j,i_3}\subseteq C_{j,i_3}$ satisfying 
$h(\Atil_{j,i_1}\times \Btil_{j,i_2}\times \Ctil_{j,i_3})\le \nu_\Omega$ and 
$$
(A_{i_1}\times B_{i_2}\times C_{i_3})\cap\Omega 
= (\Atil_{j,i_1}\times \Btil_{j,i_2}\times \Ctil_{j,i_3})\cap \Omega.
$$
Because $\calP_\Omega D_j(\bu\otimes \bv\otimes \bw)$ is supported in 
$\cup_{(i_1,i_2,i_3):i_1+i_2+i_3=j}(A_{i_1}\times B_{i_2}\times C_{i_3})\cap\Omega$, we have 
\bes
\calP_\Omega \Dtil_j(\bu\otimes \bv\otimes \bw) = \calP_\Omega D_j(\bu\otimes \bv\otimes \bw). 
\ees
This observation, together with Lemma~\ref{lm-thinning}, leads to the following characterization of the spectral norm of a tensor support on a set with bounded aspect ratio.

\begin{lemma}\label{lm-squaring} 
Let $m_\ell=\lceil \log_2 d_\ell\rceil$ for $\ell=1,2, 3$, and $D_j(\cdot)$ and $\Dtil_j(\cdot)$ be as in (\ref{D_j}) and (\ref{Dtil}) respectively. Define
\bes
\scrB^*_{\Omega,m_*} =
\Bigg\{\sum_{0\le  j\le m_*}\Dtil_j(\bu_1\otimes \bu_2\otimes \bu_3)+\sum_{m_\ast<j\le m^\ast} 
D_j(\bu_1\otimes \bu_2\otimes \bu_3): \bu_\ell\in \scrB_{m_\ell,d_\ell}\Bigg\},
\ees
and $\scrB^*_{\nu,m_*}=\cup_{\nu_{\Omega}\le\nu}\scrB^*_{\Omega,m_*}$. 
Let $\bX\in \R^{d_1\times d_2\times d_3}$ be a tensor with $\supp(\bX)\subseteq \Omega$. 
For any $0\le m_\ast\le m_1+m_2+m_3$ and $\nu\ge \nu_\Omega$, 
$$
\|\bX\| \le 8 \max_{\bY\in \scrB^*_{\Omega,m_*}} \langle \bY, \bX\rangle
\le 8 \max_{\bY\in \scrB^*_{\nu,m_*}} \langle \bY, \bX\rangle. 
$$
\end{lemma}

%%%%%%%%%%%%%%%%%%%%%%%%%%%%%

\subsection{Entropy bounds}
Essential to our argument are entropic bounds related to $\scrB^*_{\nu,m_*}$. 
It is clear that $$\#\{D_j(\bu): \|\bu\|\le 1\}\le {d_j\choose 2^k\wedge d_j}2^{2^k\wedge d_j}
\le\exp((2^k\wedge d_j)(\log 2+1+(\log(d_j/2^k))_+)),$$ so that by (\ref{class-B})
$$
|\scrB_{m_j,d_j}|\le \prod_{k=0}^{m_j} {d_j\choose 2^k\wedge d_j}2^{2^k\wedge d_j}
\le \exp\Big(d_j\sum_{\ell=1}^\infty 2^{-\ell}(\log 2+1+\log(2^\ell))\Big)
\le \exp(4.78 \,d_j)
$$
Consequently, due to $d=d_1+d_2+d_3$ and $4.78\le 21/4$, 
\bel{abs-entropy-bd}
%\#\left\{\sum_{m_\ast<j\le m^\ast}D_j(\bY): \bY\in \scrB^*_{\nu,m_*}\right\} 
\left|\scrB^*_{\nu,m_*}\right|
\le \prod_{j=1}^3 |\scrB_{m_j,d_j}|\le e^{(21/4)d}.
\eel
We derive tighter entropy bounds for slices of $\scrB^*_{\nu,m_*}$ by considering 
\bes
\scrD_{\nu,j,k} = \Big\{D_j(\bY): \bY\in \scrB^*_{\nu,m_*}, \|D_j(\bY)\|_{\rm HS}^2\le 2^{k-j} \Big\}. 
\ees
Here and in the sequel, we suppress %abbreviate 
the dependence of $\scrD$ on quantities such as $m_*,m_1,m_2,m_3$ for brevity, when no confusion occurs.

\begin{lemma}\label{lm-entropy-bd} Let $L(x,y)=\max\{1,\log(ey/x)\}$ and $\nu\ge 1$. 
%%% 
For all $0\le k\le j\le m^*$, 
\bel{lm-entropy-bd-2} 
\log\Big|\scrD_{\nu,j,k}\Big|\le (21/4)J(\nu,j,k),  
\eel
%\bes
%\log\Big|\scrD_{\nu,j,k}\Big|\le c_0 (j\vee 1) \sqrt{\nu 2^k}L\big(\sqrt{\nu 2^k},(j\vee 1)d\big). 
%\ees
where %$J(\nu,j,k)=\sqrt{\hbox{${j+2\choose 2}$}\nu 2^{k}}L\Big(\sqrt{\nu 2^k},2d\sqrt{\hbox{${j+2\choose 2}$}}\Big)$.
$J(\nu,j,k)=(j+2)\sqrt{\nu 2^{k-1}}L\big(\sqrt{\nu 2^{k-1}},(j+2)d\big)$.
\end{lemma}

\vskip 10pt
\noindent{\sc Proof of Lemma \ref{lm-entropy-bd}.} 
We first bound the entropy of a single block. Let 
\bes
\scrD^{\rm (block)}_{\nu,\ell} &=& \Big\{\sgn(u_a)\sgn(v_b)\sgn(w_c){\mathbb I}\{(a,b,c)\in A\times B\times C\}:  \cr 
&& \qquad\qquad\qquad\qquad h(A\times B\times C)\le\nu,\ |A||B||C| = \ell\Big\}. 
\ees
By the constraints on the size and aspect ratio of the block, 
\bes
\max(|A|^2,|B|^2,|C|^2) \le \nu|A||B||C|\le \nu\ell. 
\ees
By dividing $\scrD^{\rm (block)}_{\nu,\ell}$ into subsets according to $(\ell_1,\ell_2,\ell_3)=(|A|,|B|,|C|)$, we find 
\bes
\Big|\scrD^{\rm (block)}_{\nu,\ell}\Big|
\le \sum_{\ell_1\ell_2\ell_3 = \ell, \max(\ell_1,\ell_2,\ell_3)\le\sqrt{\nu\ell}} 
2^{\ell_1+\ell_2+\ell_3}{d_1\choose \ell_1}{d_2\choose \ell_2}{d_3\choose \ell_3}
\ees
By the Stirling formula, for $i=1,2,3$,
$$
\log\Big[\sqrt{2\pi\ell_i}2^{\ell_i}{d_i\choose \ell_i}\Big]
\le \ell_iL\big(\ell_i,2d\big) \le \sqrt{\nu \ell}\ L\big(\sqrt{\nu \ell},2d\big).
$$ 
%%%%
We note that $k(k+1)/(2\sqrt{q^k})$ is no greater than $2.66$, $1.16$ and $1$ respectively for 
$q=2$, $q=3$ and $q\ge 5$. 
Let $\ell = \prod_{j=1}^m q_j^{k_j}$ with distinct prime factors $q_j$. We get 
\bes
\left|\{(\ell_1,\ell_2,\ell_3): \ell_1\ell_2\ell_3=\ell\}\right| 
= \prod_{j=1}^m {k_j+1\choose 2} \le 2.66\times 1.16 \prod_{j=1}^m\sqrt{q_j^{k_j}}
\le \pi \ell^{1/2}\le\prod_{i=1}^3\sqrt{2\pi\ell_i}. 
\ees
It follows that 
\bel{lm-entropy-bd-1}
\Big|\scrD^{\rm (block)}_{\nu,\ell}\Big|\le \exp\Big(3\sqrt{\nu \ell}L\big(\sqrt{\nu \ell},2d\big)\Big). 
\eel

Due to the constraint $ i_1+i_2+i_3=j$ in defining $\scrB^*_{\nu,m_*}$, for any $\bY\in \scrB^*_{\nu,m_*}$, $D_j(\bY)$ is composed of at most $i^*= {j+2\choose 2}$ blocks. Since the sum of the sizes of the blocks is bounded by $2^k$, 
(\ref{lm-entropy-bd-1}) yields 
%%%%\bel{pf-lm-entropy-bd-1} 
\bes
\Big|\scrD_{\nu,j,k}\Big|&\le& \sum_{\ell_1+\ldots+\ell_{i^*}\le2^k} \prod_{i=1}^{i^*} \Big|\scrD^{\rm (block)}_{\nu \ell_i}\Big|\cr 
%%%%
&\le& \sum_{\ell_1+\ldots+\ell_{i^*}\le2^k} \exp\Big(\sum_{i=1}^{i^*}3\sqrt{\nu \ell_i}\ L\big(\sqrt{\nu \ell_i},2d\big)\Big)\cr
&\le& (2^k)^{i^*}\max_{\ell_1+\ldots+\ell_{i^*}\le2^k} \exp\Big(\sum_{i=1}^{i^*}3\sqrt{\nu \ell_i}\ L\big(\sqrt{\nu \ell_i},2d\big)\Big). 
\ees
It follows from the definition of $L(x,y)$ and the Cauchy-Schwarz inequality that 
\bes
\sum_{i=1}^{i^*}\sqrt{\ell_i}\ L\big(\sqrt{\nu \ell_i},2d\big)&=& \sum_{i=1}^{i^*}\sqrt{\ell_i}\Big(L\big(\sqrt{\nu 2^k},2d\big) + \log\Big(\sqrt{2^k/\ell_i}\Big)\Big)\cr 
&\le & \sqrt{2^k}\Big(\sqrt{i^*}L\big(\sqrt{\nu 2^k},2d\big)+\sum_{i=1}^{i^*}\sqrt{\ell_i/2^k}\log\Big(\sqrt{2^k/\ell_i}\Big)\Big)\cr 
&\le & \sqrt{i^*2^k}\Big(L\big(\sqrt{\nu 2^k},2d\big)+\log\big(\sqrt{i^*}\big)\Big),
\ees
where the last inequality above follows from the fact that subject to $u_1,u_2\ge 0$ and $u_1^2+u_2^2\le 2c^2\le 2$, the maximum of $-u_1\log(u_1)-u_2\log(u_2)$ is attained at $u_1=u_2=c$. 
Consequently, since $i^*\le {j+2\choose 2}$, 
\bes
\log\Big|\scrD_{\nu,j,k}\Big|\le i^*\log\big(2^{k}\big)+3\sqrt{i^*\nu 2^k}
L\left(\sqrt{\nu 2^k},2d\sqrt{\hbox{${j+2\choose 2}$}}\right). 
\ees
We note that $j\ge k$, $\sqrt{2^k}\ge k\sqrt{8}/3$, $\nu\ge 1$ and $xL(x,y)$ is increasing in $x$, so that  
\bes
\frac{\sqrt{i^*\nu 2^k}L\Big(\sqrt{\nu 2^k},2d\sqrt{\hbox{${j+2\choose 2}$}}\Big)}{i^*\log\big(2^{k}\big)}
\ge \frac{\sqrt{8}/3 L\Big(k\sqrt{8}/2,2d\sqrt{\hbox{${j+2\choose 2}$}}\Big)}{\sqrt{i^*}\log 2}
\ge \frac{\sqrt{8}\log(ed)}{\sqrt{i^*}\log 8}. 
\ees
Moreover, because $2^{m^*+3/2}\le \sqrt{8}\prod_{j=1}^3(2d_j)\le \sqrt{8}(2d/3)^3\le d^3$, 
we get $\sqrt{2i^*} \le\sqrt{(j+1)(j+2)}\le j+3/2\le  m^*+3/2\le (3/\log 2)\log d$, 
so that the right-hand side of the above inequality is no smaller than $(4/\log 8)(\log 2)/3 = 4/9$. It follows that 
\bes
\log\Big|\scrD_{\nu,j,k}\Big|\le (3+9/4)\sqrt{i^*\nu 2^k}
L\Big(\sqrt{\nu 2^k},2d\sqrt{\hbox{${j+2\choose 2}$}}\Big). 
\ees
This yields (\ref{lm-entropy-bd-2}) due to $i^*\le {j+2\choose 2}$. $\hfill\square$
\vskip 20pt

%%%%
By (\ref{abs-entropy-bd}), the entropy bound in Lemma \ref{lm-entropy-bd} is useful 
only when $0\le k\le j\le m_\ast$, where
\bel{m_*}
%%%%
m_\ast=\min\Big\{x: x\ge  m^\ast\ \hbox{ or }\ J(\nu_1,x,x)\ge d\Big\}. 
\eel

\subsection{Probability bounds}\label{subsec-prob-bd}

We are now ready to derive a useful upper bound for
$$
\max_{\bX\in\scrU(\eta)}\P\left\{\left\|{d_1d_2d_3\over n}
\sum_{i=1}^n \epsilon_i\calP_{(a_i,b_i,c_i)}\bX\right\|\ge t \right\}.
$$
%%%%
Let $\bX\in \scrU(\eta)$. For brevity, write
$\bZ_i = d_1d_2d_3\eps_i\calP_{(a_i,b_i,c_i)}\bX$ and
$$
\bZ={d_1d_2d_3\over n}\sum_{i=1}^n \epsilon_i\calP_{(a_i,b_i,c_i)}\bX={1\over n}\sum_{i=1}^n \bZ_i.
$$
Let $\Omega=\{(a_i,b_i,c_i): i\le n\}$. 
In the light of Lemma \ref{le:aspbd}, we shall proceed conditional on the event that 
$\nu_{\Omega}\le \nu_1$ in this subsection. 
In this event, Lemma \ref{lm-squaring} yields
\bel{pf-5-5-1}
\|\bZ\|&\le& 8\max_{\bY\in \scrB^*_{\nu_1,m_*}} \langle \bY, \bZ\rangle\\ \nonumber
%&=&8\max_{\bY\in \scrB^*_{\nu_1,m_*}} \sum_{j=0}^{m^\ast}\langle D_j(\bY), \bZ\rangle\\
& = & 8\max_{\bY\in \scrB^*_{\nu_1,m_*}}\left(\sum_{0\le j\le m_\ast} \langle D_j(\bY), \bZ\rangle
+\langle S_*(\bY), \bZ\rangle\right), 
\eel
where $m_*$ is as in (\ref{m_*}) with the given $\nu_1$ and $S_*(\bY) = \sum_{j>m_\ast}D_j(\bY)$. 
%We assume throughout the subsection that $n \le d_1d_2d_3/\{2e\max(d_1,d_2,d_3)\}$. 

%%%%
Let $\bY\in \scrB^*_{\nu_1,m_*}$ and $\bY_j=D_j(\bY)$. 
Recall that for $\bY_j\neq 0$, $2^{-j}\le \|\bY_j\|_{\rm HS}^2\le 1$, so that 
$\bY_j\in \cup_{k=0}^j \scrD_{\nu_1,j,k}$. 
To bound the first term on the right-hand side of (\ref{pf-5-5-1}), 
consider 
$$
\P\left\{\max_{\bY_j\in \scrD_{\nu_1,j,k}\setminus \scrD_{\nu_1,j,k-1}}\langle \bY_j,\bZ\rangle
\ge t(m_\ast+2)^{-1/2}\|\bY_j\|_{\rm HS}\right\}
$$
with $0\le k\le j\le m_*$. Because $\|\bZ_i\|_{\max}\le d_1d_2d_3\|\bX\|_{\max}\le \eta\sqrt{d_1d_2d_3}$, 
for any $\bY_j\in \scrD_{\nu_1,j,k}$,
$$
\left|\langle \bY_j, \bZ_i\rangle\right|\le 2^{-j/2}\eta\sqrt{d_1d_2d_3},\quad 
\E\langle \bY_j, \bZ_i\rangle^2\le \eta^2\|\bY_j\|_{\rm HS}^2\le \eta^2 2^{k-j}.
$$
Let $h_0(u)=(1+u)\log(1+u)-u$. By Bennet's inequality,
\bes
&& \P\left\{\langle \bY_j,\bZ\rangle \ge t2^{(k-j-1)/2}(m_\ast+2)^{-1/2}\right\}
\cr &\le& \exp\left(-{n(\eta^22^{k-j})\over 2^{-j}\eta^2d_1d_2d_3}h_0\left((t2^{(k-j-1)/2}(m_\ast+2)^{-1/2})(2^{-j/2}\eta\sqrt{d_1d_2d_3})\over \eta^22^{k-j}\right)\right)\\
&=& \exp\left(-{n2^k\over d_1d_2d_3}h_0\left(t\sqrt{d_1d_2d_3}\over \eta\sqrt{(m_\ast+2) 2^{k+1}}\right)\right). 
\ees
%%%%
Recall that $\scrD_{\nu,j,k} = \big\{\bY_j=D_j(\bY): 
\bY\in \scrB^*_{\nu,m_*}, \|\bY_j\|_{\rm HS}^2\le 2^{k-j} \big\}$. By Lemma \ref{lm-entropy-bd}, 
\bel{eq:prob-bd-0}
&&\P\left\{\max_{\bY_j\in \scrD_{\nu_1,j,k}\setminus \scrD_{\nu_1,j,k-1}}\langle \bY_j,\bZ\rangle 
\ge t(m_\ast+2)^{-1/2}\|\bY_j\|_{\rm HS}\right\}\cr
&\le&|\scrD_{\nu_1,j,k}|\max_{\bY_j\in \scrD_{\nu_1,j,k}}
\P\left\{\langle \bY_j,\bZ\rangle \ge t2^{(k-j-1)/2}(m_\ast+2)^{-1/2}\right\}\\ \nonumber
&\le& \exp\left( (21/4)J(\nu_1,j,k) -{n2^k\over d_1d_2d_3}
h_0\left(t\sqrt{d_1d_2d_3}\over \eta\sqrt{(m_\ast+2) 2^{k+1}}\right)\right).
\eel

Let $L_k=1\vee \log(ed(m_\ast+2)/\sqrt{\nu_1 2^{k-1}})$. 
By the definition of $J(\nu,j,k)$ in Lemma \ref{lm-entropy-bd}, 
\bes
J(\nu_1,j,k)\le J(\nu_1,m_\ast,k) = (m_\ast+2)\sqrt{2^{k-1}\nu_1}L_k
\le J(\nu_1,m_\ast,m_\ast) = d. 
\ees 
Let $x \ge 1$ and $t_1$ be a constant satisfying 
\bel{cond-t}
n t_1 \ge 24\eta(m_\ast+2)^{3/2}\sqrt{\nu_1d_1d_2d_3},\quad 
n  x t_1^2h_0(1) \ge 12\eta^2 (m_\ast+2)^2 \sqrt{e} d L_0. 
\eel 
We prove that for all $x\ge 1$ and $0\le k\le j\le m_*$
\bel{eq:prob-bd}
%&& 
{n2^k\over d_1d_2d_3}h_0\left(xt_1\sqrt{d_1d_2d_3}\over \eta\sqrt{(m_\ast+2) 2^{k+1}}\right)
\ge 6x J(\nu_1,m_\ast,k) \ge 6x J(\nu_1,j,k). 
\eel
Consider the following three cases: 
\bes
&& \hbox{Case 1:} \qquad \frac{xt_1\sqrt{d_1d_2d_3}}{\eta\sqrt{(m_\ast+2)2^{k+1}}} 
\ge \left[\max\left\{1,\frac{ed(m_\ast+2)}{\sqrt{\nu_1 2^{k-1}}}\right\}\right]^{1/2},
\cr && \hbox{Case 2:} \qquad 1 < \frac{xt_1\sqrt{d_1d_2d_3}}{\eta\sqrt{(m_\ast+2) 2^{k+1}}} 
\le \left[\frac{ed(m_\ast+2)}{\sqrt{\nu_1 2^{k-1}}}\right]^{1/2},
\cr && \hbox{Case 3:} \qquad \frac{xt_1\sqrt{d_1d_2d_3}}{\eta\sqrt{(m_\ast+2) 2^{k+1}}} \le 1. 
\ees
{\it Case 1:} Due to $h_0(u)\ge (u/2)\log(1+u)$ for $u\ge 1$ and the lower bound for $n t_1$, 
\bes
%&& 
{n2^k\over d_1d_2d_3}h_0\left(xt_1\sqrt{d_1d_2d_3}\over \eta\sqrt{(m_\ast+2) 2^{k+1}}\right)
&\ge& {n2^k\over d_1d_2d_3}\left(xt_1\sqrt{d_1d_2d_3}\over \eta\sqrt{(m_\ast+2)2^{k+1}}\right)\frac{L_k}{4}
\cr &\ge & 6x(m_\ast+2)\sqrt{\nu_12^{k-1}}L_k. 
\ees
{\it Case 2:} Due to $(m_\ast+2)\sqrt{2^{k-1}\nu_1}\le d$, we have 
\bes
\frac{1}{\sqrt{d_1d_2d_3}} \ge \frac{xt_1\sqrt{e}(m_\ast+2)}{\eta\sqrt{(m_\ast+2)2^{k+1}}}
\left[\frac{ed(m_\ast+2)}{\sqrt{\nu_1 2^{k-1}}}\right]^{-1}
=\frac{xt_1\sqrt{\nu_1 2^{k-1}}}{\eta d\sqrt{e(m_\ast+2)2^{k+1}}}. 
\ees
Thus, due to $h_0(u) \ge uh_0(1)$ for $u\ge 1$, we have 
\bes
{n2^k\over d_1d_2d_3}h_0\left(xt_1\sqrt{d_1d_2d_3}\over \eta\sqrt{(m_\ast+2) 2^{k+1}}\right)
&\ge& \Big(\frac{xt_1\sqrt{\nu_1 2^{k-1}}}{\eta d\sqrt{e(m_\ast+2)2^{k+1}}}\Big)
{n\sqrt{2^{k-1}} xt_1 h_0(1)\over \eta\sqrt{m_\ast+2}}
\cr &= & { nx^2 t_1^2h_0(1)\sqrt{\nu_1 2^{k-1}} \over 
2\eta^2 (m_\ast+2)\sqrt{e}d}. 
\ees
Because $n  xt_1^2h_0(1) \ge 12\eta^2 (m_\ast+2)^2 \sqrt{e} d L_0$ and $L_0\ge L_k$, it follows that  
\bes
{n2^k\over d_1d_2d_3}h_0\left(xt_1\sqrt{d_1d_2d_3}\over \eta\sqrt{(m_\ast+2) 2^{k+1}}\right)
\ge 6x (m_\ast+2)  \sqrt{\nu_12^{k-1}}L_k. 
\ees
Case 3: Due to $h_0(u) \ge u^2 h_0(1)$ for $0\le u\le 1$ and 
$n xt_1^2 h_0(1) \ge 12\eta^2 (m_\ast+2) d$, we have 
\bes
{n2^k\over d_1d_2d_3}h_0\left(xt_1\sqrt{d_1d_2d_3}\over \eta\sqrt{(m_\ast+2) 2^{k+1}}\right)
\ge {n x^2t_1^2 h_0(1) \over 2\eta^2 (m_\ast+2)} \ge 6xd \ge 6x J(\nu_1,m_*,k). 
\ees
Thus, (\ref{eq:prob-bd}) holds in all three cases. 

It follows from (\ref{eq:prob-bd-0}) and (\ref{eq:prob-bd}) that for $t_1$ satisfying (\ref{cond-t}) and all $x\ge 1$
\bes
&&\P\left\{\max_{\bY_j\in \scrD_{\nu_1,j,k}\setminus \scrD_{\nu_1,j,k-1}}\langle \bY_j,\bZ\rangle 
\ge \frac{xt_1\|\bY_j\|_{\rm HS}}{\sqrt{m_\ast+2}}\right\}
\le \exp\left(-(6x-21/4)J(\nu_1,m_*,k)\right). 
\ees
We note that $J(\nu_1,m_\ast,k)\ge J(1,m_\ast,0)\ge (m_\ast+2)\log(ed(m_\ast+2))$
by the monotonicity of $x\log(y/x)$ for $x\in [1,y]$ and $\sqrt{\nu_12^{k-1}}\ge 1$. 
Summing over $0\le k\le j\le m_\ast$, we find by the union bound that 
\bes
&& \P\left\{\max_{0\le j\le m_\ast}\max_{\bY\in \scrB^\ast_{\nu_1,m_\ast}: 
D_j(\bY)\neq 0}{\langle D_j(\bY),\bZ\rangle\over \|D_j(\bY)\|_{\rm HS}}
\ge \frac{xt_1}{(m_\ast+2)^{1/2}}\right\}
\cr &\le& {m_*+2\choose 2}\{ed(m_\ast+2)\}^{-(6x-21/4)(m_\ast+2)}
\ees

%%%%
For the second term on the right-hand side of (\ref{pf-5-5-1}), we have 
\bes
\left|\langle S_*(\bY) , \bZ_i\rangle\right| 
\le 2^{-m_\ast/2}\eta \sqrt{d_1d_2d_3},\quad 
\E\left(\langle S_*(\bY) , \bZ_i\rangle\right)^2
\le \eta^2\|S_*(\bY)\|_{\rm HS}^2.
\ees
As in the proof of (\ref{eq:prob-bd-0}) and (\ref{eq:prob-bd}), 
$\log |\scrB^*_{\nu_1,m_*}|\le 5d = 5J(\nu_1,m_\ast,m_\ast)$ implies 
%%%%
\bes
&& \P\left\{\max_{\bY\in \scrB^*_{\nu_1,m_*}: 
S_*(\bY)\neq 0}{\langle S_*(\bY),\bZ\rangle \over \|S_*(\bY)\|_{\rm HS}}
\ge \frac{xt_1}{(m_\ast+2)^{1/2}}\right\}
\cr &\le& (m^\ast-m_\ast)\{ed(m_\ast+2)\}^{-(6x-21/4)(m_\ast+2)}
\ees
By Cauchy Schwartz inequality, for any $\bY\in \scrB^*_{\nu_1,m_*}$,
$$
\frac{\|S_*(\bY)\|_{\rm HS}+\sum_{0\le j\le m_\ast} \|D_j(\bY)\|_{\rm HS}}{(m_\ast+2)^{1/2}} 
\le \left(\|S_*(\bY)\|_{\rm HS}^2+\sum_{0\le j\le m_\ast} \|D_j(\bY)\|_{\rm HS}^2\right)^{1/2}\le 1. 
$$
Thus, by (\ref{pf-5-5-1}), for all $t_1$ satisfying (\ref{cond-t}) and $x\ge 1$, 
\bel{prob-bd-2}
&& \P\left\{\max_{\bY\in \scrB^*_{\nu_1,m_*}}\langle \bY, \bZ\rangle\ge xt_1\right\}
\cr &\le& \left\{{m_*+2\choose 2}+m^\ast-m_\ast\right\}\{ed(m_\ast+2)\}^{-(6x-21/4)(m_\ast+2)}.
\eel
We now have the following probabilistic bound via (\ref{prob-bd-2}) and Lemma \ref{le:aspbd}. 

\begin{lemma}\label{lm-prob-1} 
Let $\nu_1$ be as in Lemma \ref{le:aspbd}, $x\ge 1$, 
$t_1$ as in (\ref{cond-t}) and $m_\ast$ in (\ref{m_*}). Then,  
\bes
&& \max_{\bX\in\scrU(\eta)}
\P\left\{\left\|{d_1d_2d_3\over n}\sum_{i=1}^n \epsilon_i\calP_{(a_i,b_i,c_i)}\bX\right\|
\ge xt_1\right\}
\cr && \le \left\{{m_*+2\choose 2}+m^\ast-m_\ast\right\}\{ed(m_\ast+2)\}^{-(6x-21/4)(m_\ast+2)}+d^{-\beta-1}/3. 
\ees
\end{lemma}

\subsection{Proof of Lemma \ref{lm-tensor-sparse-ineq}}
We are now in position to prove Lemma \ref{lm-tensor-sparse-ineq}. 
Let $\Omega_1=\{(a_i,b_i,c_i), i\le n_1\}$. 
By the definition of the tensor coherence $\alpha(\bX)$ and the conditions on $\alpha(\bT)$ and $\rbar(\bT)$, 
we have $\|\bW\|_{\max}\le \eta/\sqrt{d_1d_2d_3}$ with $\eta = (\alpha_0\sqrt{r})\wedge\sqrt{d_1d_2d_3}$, so that 
in the light of Lemmas \ref{lm-sym}, \ref{le:aspbd} and \ref{lm-prob-1}, 
\bel{new-1}
&&\max_{\substack{\bX: \bX=\calQ_{\bT}\bX\\ \|\bX\|_{\max}\le \|\bW\|_{\max}}}\P\left\{\left\|{d_1d_2d_3\over n_1}\sum_{i=1}^{n_1} \calP_{(a_i,b_i,c_i)}\bX-\bX\right\|\ge {1\over 8}\right\}\\ \nonumber
&\le & 4\exp\left(-{{n_1(1/16)^2/2}\over \eta^2+(2/3)\eta(1/16)\sqrt{d_1d_2d_3}}\right)
\cr && +\left\{{m_*+2\choose 2}+m^\ast-m_\ast\right\}\{ed(m_\ast+2)\}^{-(6x-21/4)(m_\ast+2)}+d^{-\beta-1}/3
\eel
with $t=1/8$ in Lemma \ref{lm-sym} and the $\nu_1$ in Lemma \ref{le:aspbd}, provided that 
$xt_1\le 1/16$ and 
\bes
n_1 t_1 \ge 24\eta(m_\ast+2)^{3/2}\sqrt{\nu_1d_1d_2d_3},
\quad n_1 xt_1^2h_0(1) \ge 12\eta^2 (m_\ast+2)^2 \sqrt{e} d L_0. 
\ees
Thus, the right-hand side of (\ref{new-1}) is no greater than 
$d^{-\beta-1}$ for certain $x>1$ and $t_1$ satisfying these conditions when 
\bes
n_1 &\ge& c_1'\Big(1+\frac{1+\beta}{m_\ast+2}\Big)
\Big(\eta\sqrt{(m_\ast+2)^3\nu_1d_1d_2d_3}+\eta^2(m_\ast+2)^2 L_0d\Big)
\cr && + c_1'(1+\beta)(\log d)\Big(\eta^2 + \eta\sqrt{d_1d_2d_3}\Big)
\ees
for a sufficiently large constant $c_1'$. 
Because $\sqrt{2^{m_\ast-1}}\le J(\nu_1,m_\ast,m_\ast) = d$, it suffices to have
\bes
n_1 \ge c_1''\left[\big(\beta + \log d\big)
\Big\{\eta\sqrt{(\log d)\nu_1d_1d_2d_3}+\eta^2(\log d)^2 d\Big\}
+(1+\beta)(\log d)\eta\sqrt{d_1d_2d_3}\right]. 
\ees
When the sample size is $n_1$, 
$\nu_1 = (d^{\delta_1} en_1\max(d_1,d_2,d_3)/(d_1d_2d_3))\vee \{(3+\beta)/\delta_1\}$ in Lemma \ref{le:aspbd}
with $\delta_1\in [1/\log d,1]$. 
When $\nu_1=d^{\delta_1} en_1\max(d_1,d_2,d_3)/(d_1d_2d_3)$, 
$n_1 \ge x \sqrt{\nu_1d_1d_2d_3}$ iff 
$n_1 \ge x^2 d^{\delta_1} e\max(d_1,d_2,d_3)$. 
Thus, it suffices to have 
\bes
n_1 &\ge& c_1''\left[\big(\beta + \log d\big)^2\eta^2(\log d)d^{1+\delta_1}
+ \big(\beta + \log d\big)\eta\sqrt{(\log d)(3+\beta)\delta_1^{-1}d_1d_2d_3}\right]
\cr && + c_1''\left[\big(\beta + \log d\big)\eta^2(\log d)^2 d
+(1+\beta)(\log d)\eta\sqrt{d_1d_2d_3}\right]. 
\ees
Due to $\sqrt{(1+\beta)(\log d)}\le 1+\beta+\log d$, the quantities in the second line in the above inequality 
is absorbed into those in the first line. Consequently, with $\eta = \alpha_0\sqrt{r}$, 
the stated sample size is sufficient. $\hfill\square$

\section{Discussions}
\label{sec:disc}

In this paper, we study the performance of nuclear norm minimization in recovering a large tensor with low Tucker ranks. Our results demonstrate the benefits of not treating tensors as matrices despite its popularity.

Throughout the paper, we have focused primarily on third order tensors. In principle, our technique can also be used to treat higher order tensors although the analysis is much more tedious and the results quickly become hard to describe. Here we outline a considerably simpler strategy which yields similar sample size requirement as the vanilla nuclear norm minimization. The goal is to illustrate some unique and interesting phenomena associated with higher order tensors. 

The idea is similar to matricization -- instead of unfolding a $N$th order tensor into a matrix, we unfold it into a cubic or nearly cubic third order tensor. To fix ideas, we shall restrict our attention to hyper cubic $N$th order tensors with $d_1=\ldots=d_N=:d$ and $r_1(\bT), r_2(\bT), \ldots, r_N(\bT)$ are bounded from above by a constant. The discussion can be straightforwardly extended to more general situations. In this case, the resulting third order tensor will have dimensions either $d^{\lfloor N/3\rfloor}$ or $d^{\lfloor N/3\rfloor+1}$, and Tucker ranks again bounded. Here $\lfloor x\rfloor$ stands for the integer part of $x$. 
%It is not hard to see that if the original $N$th order tensor is incoherent in that both
%$$
%\mu(\bX)=\max_{1\le j\le N} \mu(\calL_j(\bX)),\qquad {\rm and}\qquad \alpha(\bX)=d^{N/2}r^{-1/2}\|\bW\|_{\max},
%$$
%where $\calL_j(\bX)$ is the linear space spanned by the $j$th fiber of $\bX$, and $\bW$ is an $N$th order tensor such that $\calL_j(\bW)\subseteq \calL_j(\bX)$, $\|\bW\|=1$ and $\langle \bW, \bX\rangle =\|\bX\|_\ast$, are bounded, so is the unfolded third order tensor. 
%Now, we can apply the nuclear norm minimization technique to recover the unfolded third order tensor. 
Our results on third order tensor then suggests a sample size requirement of
$$
n\asymp d^{N/2}\polylog(d).
%|\Omega|\gg \left\{\begin{array}{ll}r^{2N/3}d^{N/3}+r^{N/6}d^{N/2}& {\rm if\ } N {\rm \ mod\ }3=0\\
%r^{2\lfloor N/3\rfloor}d^{\lfloor N/3\rfloor+1}+r^{(\lfloor N/3\rfloor+1)/2}d^{N/2}& {\rm if\ } N {\rm \ mod\ }3=1\\
%r^{2\lfloor N/3\rfloor+1}d^{\lfloor N/3\rfloor+1}+r^{\lfloor N/3\rfloor/2+1}d^{N/2}& {\rm if\ } N {\rm \ mod\ }3=2
%\end{array}\right..
$$
%In particular, when $r$ is also bounded, the sample size requirement can be simplified as $|\Omega|\gg d^{N/2}$ which can be shown to be identical to that for directly applying the nuclear norm minimization to the original $N$th order tensor. 
This is to be compared with a matricization approach that unfolds an $N$th order tensor to a (nearly) square matrix (see, e.g., Mu et al., 2013) -- where %again assuming that $r$ is bounded, 
the sample size requirement is $d^{\lceil N/2\rceil}\polylog(d)$. It is interesting to notice that, in this special case, unfolding a higher order tensor to a third order tensor is preferable to matricization when $N$ is odd.

\end{document}